\documentclass{article}

\usepackage[final]{corl_2026} 
\usepackage{subfiles}
\usepackage{amsmath}
\usepackage{hyperref}
\usepackage{enumitem}
\usepackage{subfiles}
\usepackage{multirow}
\usepackage{booktabs}
\usepackage{graphics}
\usepackage{graphicx}
\usepackage{amsmath}
\usepackage{amssymb}
\usepackage{xcolor}
\usepackage{pifont}
\usepackage{booktabs}
\usepackage{graphicx}

\usepackage{tikz}
\usetikzlibrary{arrows.meta}
\usepackage{xcolor}

\definecolor{affinecolor}{RGB}{55,175,215}      
\definecolor{activationcolor}{RGB}{180,180,180} 
\definecolor{prismcolor}{RGB}{245,140,32}       
\definecolor{prismtextcolor}{RGB}{190,95,0}     

\tikzset{
  conn/.style={
    -{Latex[length=1.0mm,width=.75mm]},
    thin, shorten <=.8pt, shorten >=.8pt
  }
}

\usepackage[table]{xcolor}

\newcommand{\sname}{PRISM}
\definecolor{cmarkgreen}{RGB}{0,150,80}
\definecolor{xmarkred}{RGB}{200,50,50}

\newcommand{\cmark}{\textcolor{cmarkgreen}{\ding{51}}}
\newcommand{\xmark}{\textcolor{xmarkred}{\ding{55}}}
\usepackage[most]{tcolorbox}

\newtcolorbox{rqbox}{
    colback=gray!5,
    colframe=gray!55,
    boxrule=0.6pt,
    arc=2pt,
    left=6pt,
    right=6pt,
    top=5pt,
    bottom=5pt
}

\title{\sname: Polynomial Representations for Interaction-Structured Motor Control}

\author{
  Seung Hyun Lee \hspace{8mm} Stella X. Yu\\[2pt]
  University of Michigan\\
}

\usepackage{multirow}
\begin{document}

\maketitle

\begin{abstract}
Robot policies map observations to actions, typically using multilayer perceptrons (MLPs). Yet robot dynamics and control objectives involve multiplicative interactions among observable physical variables.  MLPs can approximate these interactions but must learn them implicitly, at a representational cost. We introduce PRISM (\underline{P}olynomial \underline{R}epresentations for \underline{I}nteraction-\underline{S}tructured Motor Control), a physics-motivated yet data-driven representation that learns multiplicative interactions by recursively adding scaled bilinear features, without enumerating all polynomial terms.  Replacing only the MLP's first affine layer, PRISM is plug-and-play for reinforcement and imitation learning, requiring no changes to training objectives, sensing, or controllers.  Across humanoid locomotion and contact-rich manipulation, PRISM improves survival from 51.0\% to 92.5\%, outperforming even larger MLPs. More strikingly, its performance advantage persists under out-of-distribution changes in friction,
robot mass, and payload, without retraining. PRISM also exhibits sensorless compliance-like behavior, increases failure-recovery success from
40.9\% to 52.3\% without explicit
admittance control, improves real-robot manipulation, and learns features that better predict physical quantities.  These results make PRISM a compelling alternative to standard MLPs for robot policy learning. The project page is available at~\url{https://lsh3163.github.io/prism/}.
\end{abstract}

\keywords{Polynomial interaction, Motor control, Sensorless compliance}

\def\figKeyDiagram#1{%
\begin{figure}[#1]
\centering
\setlength{\tabcolsep}{26pt}%
\begin{tabular}{@{}cc@{}}
\begin{tikzpicture}[
  x=3cm,y=3cm,
  baseline={(current bounding box.center)}
]
  \path[use as bounding box]
    (0.03,-.225) rectangle (1.15,.23);

  \draw[thin]
    (0.04,.22) rectangle (1.14,-.215);

  \draw[affinecolor,line width=2.9pt]
    (.18,-.18)--(.18,.17);
  \draw[activationcolor,line width=2.9pt]
    (.28,-.18)--(.28,.17);

  \draw[affinecolor,line width=2.9pt]
    (.58,-.18)--(.58,.04);
  \draw[activationcolor,line width=2.9pt]
    (.68,-.18)--(.68,.04);

  \draw[affinecolor,line width=2.9pt]
    (.90,-.18)--(.90,-.045);
  \draw[activationcolor,line width=2.9pt]
    (1.00,-.18)--(1.00,-.045);

  \node[anchor=south east,font=\scriptsize]
    at (1.07,.075) {MLP};
\end{tikzpicture}
&
\begin{tikzpicture}[
  x=3cm,y=3cm,
  baseline={(current bounding box.center)}
]
  \path[use as bounding box]
     (-.13,-.225) rectangle (1.83,.23);

  \draw[thin]
    (-.12,.22) rectangle (1.82,-.215);

  \filldraw[
    fill=prismcolor!10,
    draw=prismcolor,thin
  ]
    (-.01,.175) rectangle (.86,-.175);

  \draw[affinecolor,line width=2.9pt]
    (.20,.035)--(.20,.15);
  \draw[affinecolor,line width=2.9pt]
    (.20,-.15)--(.20,-.035);

  \node[anchor=east,font=\scriptsize]
    at (.21,.095) {$A_1$};
  \node[anchor=east,font=\scriptsize]
    at (.21,-.095) {$A_2$};

  \node[
    draw,circle,minimum size=2.7mm,
    inner sep=0pt,font=\scriptsize,fill=white
  ] (mul) at (.47,-.095) {$\cdot$};

  \node[
    draw,minimum size=2.7mm,
    inner sep=0pt,font=\scriptsize,fill=white
  ] (add) at (.73,.095) {$+$};

  \draw[conn] (.20,.095)--(add.west);
  \draw[conn] (.20,-.095)--(mul.west);
  \draw[conn] (.20,.095)--([yshift=1.0pt]mul.west);

  \draw[conn] (mul.east) --
    node[above=.2pt,pos=.48,font=\scriptsize] {$s_2 {\scriptsize\times}$}
    (.73,-.095) -- (add.south);

  \draw[activationcolor,line width=2.9pt]
    (.96,-.18)--(.96,.17);

  \draw[affinecolor,line width=2.9pt]
    (1.26,-.18)--(1.26,.04);
  \draw[activationcolor,line width=2.9pt]
    (1.36,-.18)--(1.36,.04);

  \draw[affinecolor,line width=2.9pt]
    (1.58,-.18)--(1.58,-.045);
  \draw[activationcolor,line width=2.9pt]
    (1.68,-.18)--(1.68,-.045);

  \node[anchor=south east,font=\scriptsize]
    at (1.75,.075) {PRISM};
\end{tikzpicture}
\end{tabular}

\caption{
\textbf{\sname{} adds learnable polynomial interactions to MLPs by replacing only the first affine layer.}
{\bf Left}: A three-layer MLP with \textcolor{affinecolor}{affine} and \textcolor{activationcolor}{ELU} layers.
{\bf Right}: The orange-shaded block in \textcolor{prismcolor}{PRISM} combines two affine branches as $A_1+s_2\times(A_1\odot A_2)$. Additional branches can recursively extend the representation to higher orders; all other layers remain unchanged.
}
\label{fig:keyDiagram}
\end{figure}%
}

\def\PRKeyResultsChart{%
\begingroup%
\def\PRPlotHeight{27}
\def\PRPlotBottom{29}%
\def\PRBarWidth{5.0}
\def\PRBarGap{1.4}
\def\PRValueSize{7.0}%
\def\PRTickSize{7.0}%
\def\PRTitleSize{8.0}%
\def\PRDecimals{1}%
\def\PRCategoryY{25.5}%
\def\PRBaselineRowY{15.1}%
\def\PRMethodRowY{6.4}%

\colorlet{PRBaseline}{affinecolor}%
\colorlet{PRMethod}{prismcolor}%
\colorlet{PRMethodText}{prismtextcolor}%
\definecolor{PRText}{HTML}{222222}%
\definecolor{PRAxis}{HTML}{888888}%
\definecolor{PRGrid}{HTML}{E3E3E3}%


\def\PRPolicyData{45/51.0/92.5,91/63.5/67.6,137/62.0/63.82}%

\def\PRPracticalData{190/40.9/52.3,245/68.0/94.0}%

\def\PRDynamicsData{310/78.3/92.3,367/80.0/87.1}%

\pgfmathsetlengthmacro{\PRXUnit}{0.96*\textwidth/396}%
\pgfmathsetmacro{\PRTop}{\PRPlotBottom+\PRPlotHeight}%
\pgfmathsetmacro{\PRTitleY}{\PRTop+8.3}%
\pgfmathsetmacro{\PRCanvasHeight}{\PRTop+11}%

\begin{tikzpicture}[
  x=\PRXUnit,y=1bp,
  every node/.style={inner sep=0pt,outer sep=0pt,text=PRText},
  pr title/.style={font=\fontfamily{phv}\selectfont\fontsize{\PRTitleSize bp}{9bp}\selectfont},
  pr tick/.style={font=\fontfamily{phv}\selectfont\fontsize{\PRTickSize bp}{8bp}\selectfont},
  pr value/.style={font=\fontfamily{phv}\selectfont\fontsize{\PRValueSize bp}{8bp}\selectfont},
  pr bar/.style={draw=none}
]%
\path[use as bounding box] (0,0) rectangle (396,\PRCanvasHeight);%

\def\PRPanel##1##2##3{%
  \foreach \v in {50,100}{%
    \draw[PRGrid,line width=0.35bp]
      (##1,{\PRPlotBottom+\PRPlotHeight*\v/100}) --
      (##2,{\PRPlotBottom+\PRPlotHeight*\v/100});%
  }%
  \draw[PRAxis,line width=0.45bp]
    (##1,\PRTop) -- (##1,\PRPlotBottom) -- (##2,\PRPlotBottom);%
  \node[pr title,anchor=center] at ({(##1+##2)/2},\PRTitleY) {##3};%
}%

\PRPanel{25}{156}{task performance}%
\PRPanel{167}{268}{practical robot control}%
\PRPanel{281}{395}{dynamics generalization}%

\foreach \v in {0,50,100}{%
  \draw[PRAxis,line width=0.45bp]
    (23.5,{\PRPlotBottom+\PRPlotHeight*\v/100}) --
    (25,{\PRPlotBottom+\PRPlotHeight*\v/100});%
  \node[pr tick,anchor=east]
    at (20.5,{\PRPlotBottom+\PRPlotHeight*\v/100}) {\v\%};%
}%

\def\PRNumber##1{%
  \pgfmathprintnumber[fixed,fixed zerofill,precision=\PRDecimals,
    assume math mode=true]{##1}%
}%

\def\PRPair##1##2##3{%
  \pgfmathsetmacro{\PRBaseY}{\PRPlotBottom+\PRPlotHeight*(##2)/100}%
  \pgfmathsetmacro{\PRMethodY}{\PRPlotBottom+\PRPlotHeight*(##3)/100}%
  \path[pr bar,fill=PRBaseline]
    ({##1-\PRBarGap/2-\PRBarWidth},\PRPlotBottom)
    rectangle ({##1-\PRBarGap/2},\PRBaseY);%
  \path[pr bar,fill=PRMethod]
    ({##1+\PRBarGap/2},\PRPlotBottom)
    rectangle ({##1+\PRBarGap/2+\PRBarWidth},\PRMethodY);%
  \node[pr value,text=PRBaseline,anchor=center]
    at (##1,\PRBaselineRowY) {\PRNumber{##2}};%
  \node[pr value,text=PRMethodText,anchor=center]
    at (##1,\PRMethodRowY) {\PRNumber{##3}};%
}%

\foreach \x/\b/\p in \PRPolicyData
  {\PRPair{\x}{\b}{\p}}%
\foreach \x/\b/\p in \PRPracticalData
  {\PRPair{\x}{\b}{\p}}%
\foreach \x/\b/\p in \PRDynamicsData
  {\PRPair{\x}{\b}{\p}}%

\foreach \x/\lab in {
  45/Walking,
  91/Manipulation,
  137/Multi-task,
  190/Drop~recovery,
  245/Tape~pickup,
  310/Flat~(60\,s),
  367/Rough~(20\,s)
}{%
  \node[pr tick,anchor=north]
    at (\x,\PRCategoryY) {\lab};%
}%

\node[pr value,text=PRBaseline,anchor=west]
  at (0,\PRBaselineRowY) {Base};%
\node[pr value,text=PRMethodText,anchor=west]
  at (0,\PRMethodRowY) {PRISM};%

\end{tikzpicture}%
\endgroup%
}%

\def\figKeyResults#1{%
\begin{figure}[#1]
\centering
\PRKeyResultsChart
\vspace{-0.7em}
\caption{%
\textbf{\sname{} improves task performance and practical robot control,
while generalizing to out-of-distribution physical conditions.}
Cyan and orange denote baseline and \sname{} policies.
\emph{Left:} Humanoid-Gym~\cite{gu2024humanoid} survival
and manipulation success on
ManiSkill3~\cite{tao2025maniskill3} and
LIBERO~\cite{liu2023libero}.
\emph{Middle:} StackCube~\cite{tao2025maniskill3}
failure recovery and real-robot SO-101 tape pickup.
\emph{Right:} Mean survival rates across six out-of-range
dynamics settings on each terrain: flat (60-second trials)
and rough (20-second trials), varying friction, mass, and payload.  \sname{} outperforms baselines across settings.
}
\label{fig:keyResults}
\end{figure}
}%

\section{Introduction}

Every robot policy answers the same question: Given what the robot can observe, what action should it take next? In learning-based robot control, multilayer perceptrons (MLPs) are widely used to map observations to actions.  MLPs
are general-purpose function approximators that can learn mappings between diverse inputs and outputs without assuming how they are related. This flexibility suits visual classification, where low-level light intensities are mapped to high-level semantic categories.

Robot control, however, involves a different input-output relationship. In visual classification, an output label describes the input image but does not change it. A robot policy, by contrast, acts on the same physical body it observes: Its action outputs change the robot's state, which in turn changes its subsequent proprioceptive inputs. This input-output coupling is governed by the body's dynamics, even though these dynamics are not explicitly provided to the policy. For articulated robots, Newtonian dynamics take the form~\cite{lynch2017modern}:
\begin{equation}
M(q)\ddot q + C(q,\dot q)\dot q + g(q) = Q,
\label{eq:dynamics}
\end{equation}
where $q$ describes the robot configuration, including joint positions, $\dot q$ the corresponding velocities, $M$ the mass matrix, $C(q,\dot q)\dot q$ the Coriolis and centrifugal effects, $g(q)$ the generalized gravitational forces and torques, and $Q$ the applied generalized forces. Crucially, these dynamics couple joint motions through products of physical variables.

Multiplicative interactions extend beyond robot dynamics to control objectives. Energy-efficient motion and regulated contact involve quantities such as joint mechanical power and contact work. These relationships share a common algebraic structure -- products of physical variables:
\begin{equation}
[C(q,\dot q)\dot q]_i
= \sum_{j,k}c_{ijk}(q)\dot q_j\dot q_k,
\qquad
P_{\mathrm{joint}} = \sum_i \tau_i\dot q_i,
\qquad
W_{\mathrm{contact}} =
\int_{t_0}^{t_1} f_c^\top v_c\,dt.
\end{equation}
Although some variables such as joint torques and contact forces may not be directly observed, their effects can be reflected in proprioceptive and action histories. Learning effective actions can thus benefit from explicitly representing multiplicative interactions among available observations.

Standard MLPs, however, do not represent such interactions directly. They compose affine layers with nonlinear activations such as ReLU~\cite{nair2010rectified} or ELU~\cite{clevert2016fast}. They can approximate multiplicative interactions~\cite{cybenko1989approximation,hornik1989multilayer}, but must learn them implicitly. Finite ReLU networks are piecewise affine and cannot represent quadratic products exactly; greater accuracy requires more network complexity~\cite{yarotsky2017error}. Smoother activations such as ELU avoid this restriction, but still learn these interactions implicitly, consuming capacity in the shallow MLPs common in robot control.

We propose \textbf{\sname{}}, \textbf{P}olynomial \textbf{R}epresentations for \textbf{I}nteraction-\textbf{S}tructured \textbf{M}otor Control, to explicitly represent and learn multiplicative interactions among policy observations (Fig.~\ref{fig:keyDiagram}).
{\bf 1) We learn multiplicative interactions rather than prescribe them.}
Rather than specifying robot dynamics or handcrafting physical quantities, \sname{} learns useful products among task commands, proprioceptive state histories, and past actions, providing a physics-motivated inductive bias without imposing a dynamics model.
{\bf 2) We represent these interactions compactly by recursively adding bilinear terms.}
Rather than enumerating polynomial terms, \sname{} constructs degree-$K$ features from $K$ affine branches through scaled element-wise products. Replacing only the first affine layer of an MLP, \sname{} is plug-and-play for reinforcement and imitation learning, requiring no changes to training objectives, controllers, or sensing. 
Zeroing all scales reduces \sname{} to a standard MLP.

\figKeyDiagram{h}

\figKeyResults{h}

Experiments on humanoid locomotion and contact-rich manipulation reveal several benefits (Fig.~\ref{fig:keyResults}).
{\bf 1)} \sname{} substantially outperforms standard MLPs, raising survival in Humanoid-Gym~\cite{gu2024humanoid} from 51.0\% to 92.5\%, gains not matched by larger MLPs. 
{\bf 2)} It exhibits sensorless compliance-like behavior, rapidly relaxing contact force without explicit admittance
control.  It also increases failure-recovery success from 40.9\% to 52.3\%. On a physical SO-101, tape-pickup success rises from 68\% to 94\%.
{\bf 3)} More strikingly, its advantage persists under out-of-distribution shifts in friction, mass, and payload, without retraining.
{\bf 4)} Linear probes show that its features better predict mechanics-inspired quantities, including joint power, slip, contact impulse, and a contact-work proxy.
Together, these findings make \sname{} a compelling alternative to standard MLPs in robot policy learning.
\section{Related Work}
\label{sec:related}

\paragraph{Polynomial and multiplicative neural networks.}
Multiplicative interactions expand the representable function classes of
neural networks and provide a useful inductive bias for coupling multiple
information streams~\cite{jayakumar2020multiplicative}.
Polynomial neural networks further use factorized parameterizations to
represent higher-order dependencies without explicitly enumerating all
monomials~\cite{chrysos2021deep,usevich2025identifiability,maharajdeep}.
Polynomial expansions have also been studied as a tool for analyzing the
behavior of trained neural networks~\cite{xiao2024hope}. Related work in robot learning incorporates physical priors through auxiliary
estimators, constrained objectives, or physics-informed loss
functions~\cite{li2025physics,hu2026quietwalk,guo2025physics}.
These approaches explicitly estimate selected physical quantities or enforce
task-specific consistency constraints during training. In contrast,
\sname{} does not impose a physical model or require auxiliary physical
supervision. Instead, it introduces a factorized low-degree interaction structure into the proprioceptive representation itself. Thus \sname{} learns physical interactions from data through a structural inductive bias, while remaining sensor-minimal at deployment. 

\paragraph{Robotic control with proprioceptive data.}
Practical deployment often imposes strict sensing constraints, making
proprioception central to robust robot control~\cite{peng2020learning,
lee2020learning,zhang2024learning,zhi2025learning}.
Legged-robot policies commonly use privileged training and history-based
adaptation modules to infer the effects of changing terrain, payload, and
dynamics~\cite{kumar2021rma}. Recent humanoid systems additionally emphasize
integrated perceptual-motor representations for complex terrain and fall
recovery~\cite{azulay2026vigor,lin2025let}. In contact-rich manipulation, other approaches introduce additional
interaction modalities, including force, tactile, audio, or contact-related
observations~\cite{zhang2026glove2hand,liu2024maniwav,wang2025sound,
zhu2026touch,jiang2025kaiwu}, or employ specialized compliance
controllers~\cite{xu2025facet,margolis2025softmimic,shi2026minimalist}.
\sname{} is complementary to these approaches. Rather than reconstructing unobserved physical parameters, it exposes interactions among signals already available to the policy. This allows the policy to represent the observable consequences of changing contact and dynamics directly within its proprioceptive conditioning pathway.

\section{Polynomial Representations for Interaction-Structured Motor Control}
\label{sec:method}

\begin{figure*}[!b]
\centering
\includegraphics[width=\textwidth]{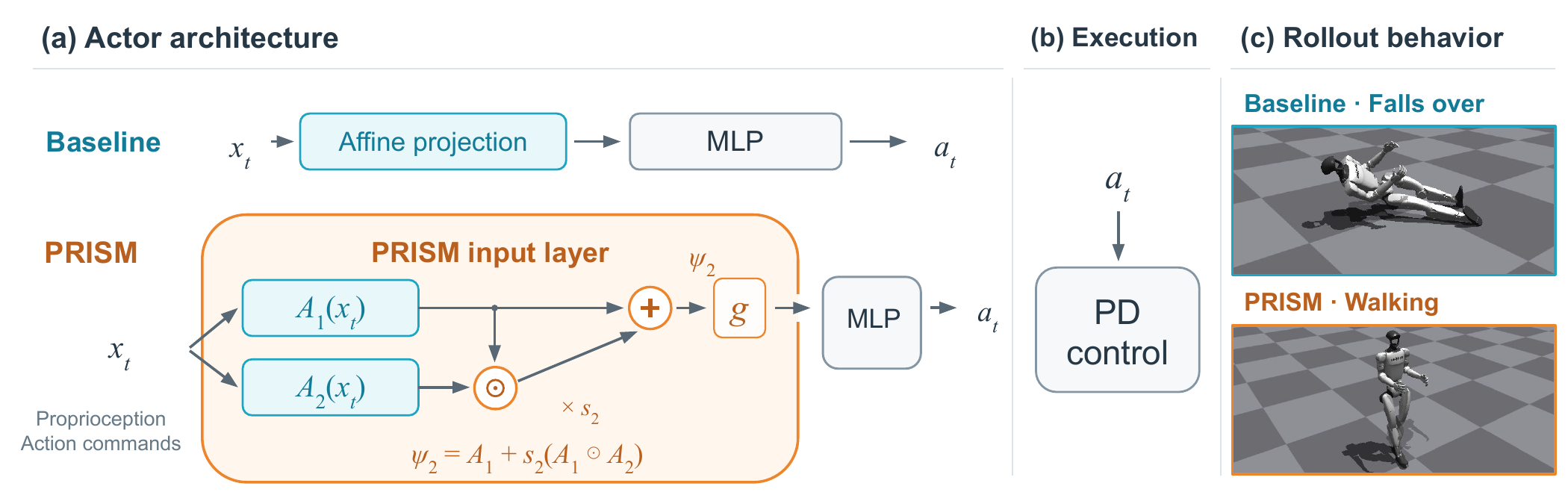}
\caption{
\textbf{\sname{} replaces the initial affine projection
with a factorized polynomial representation while retaining the downstream architecture.}
(a) Actor architecture.
The baseline uses an affine projection followed by an MLP.
\sname{} instead combines two learned affine factors,
$A_1(x_t)=W_1x_t+b_1$ and $A_2(x_t)=W_2x_t+b_2$,
to construct
$\psi_2(x_t)=A_1(x_t)+s_2(A_1(x_t)\odot A_2(x_t))$.
The projection $g$ maps the resulting features
to the downstream MLP, whose architecture remains unchanged.
(b) Execution.
Both policies output actions through the same
low-level PD controller.
(c) Rollout behavior. Rollouts show the baseline
falling (top) and \sname{} walking (bottom).
}
\vspace{-0.5em}
\label{fig:method_prism}
\end{figure*}

\sname{} introduces a factorized polynomial representation
into the observed-state pathway of a robot policy.
It augments a direct affine path with products of learned
affine factors, making multiplicative interactions
available to the downstream policy.
Fig.~\ref{fig:method_prism} illustrates the representation
and its integration into reinforcement-learning actors
and imitation-learning conditioning pathways.
We first describe the policy inputs, then present the
polynomial construction and its use in policy learning.


\paragraph{Policy inputs.}
Let $\pi_\theta(a_t \mid o_t)$ denote a policy that
predicts action $a_t$ from input $o_t$.
We partition the input as $o_t=(x_t,c_t)$, where
$x_t \in \mathbb{R}^{d}$ contains the state, command,
and history inputs processed by \sname{}, and $c_t$
contains the remaining policy inputs.
\begin{itemize}[leftmargin=*,itemsep=1pt,topsep=1pt]
    \item \textbf{Locomotion.}
    $x_t$ contains the actor's proprioceptive observations
    together with commands and observation history.
    Proprioceptive signals describe the robot's configuration
    and motion, such as joint positions and velocities,
    base angular velocity, and projected gravity.
    Commands and previous actions, when included in the
    actor input, participate in the learned interactions
    alongside the observed state.
    The observation channels and temporal sampling follow
    the original policy.
    
    \item\textbf{Manipulation.}
    $x_t$ contains the proprioceptive input or history,
    while $c_t$ contains RGB observations and, when available,
    language instructions.
    \sname{} processes the proprioceptive pathway;
    visual and language inputs retain their existing
    encoders and conditioning pathways.
    The resulting polynomial features are combined with
    these modalities for action prediction.
\end{itemize}

\paragraph{Factorized polynomial representation.}
For polynomial degree $K$, we introduce $K$ affine feature layers, each with weight matrix $W_k$ and bias $b_k$:
\begin{equation}
A_k(x_t) = W_k x_t + b_k, \qquad k=1,\ldots,K.
\end{equation}
\sname{} constructs polynomial features by recursively adding scaled multiplicative interactions:
\begin{align}
&\text{Initialization: }&
\psi_1(x_t) &= A_1(x_t),\\
&\text{For $k=2$: }&
\psi_2(x_t) &= A_1(x_t) +
s_2\left(A_1(x_t)\odot A_2(x_t)\right),
\label{eq:prism_degree2}\\
&\text{For $k\ge 2$: }&
\psi_k(x_t) &= \psi_{k-1}(x_t) +
s_k\left(\psi_{k-1}(x_t)\odot A_k(x_t)\right).
\label{eq:prism_recursive}
\end{align}
Here, $\odot$ denotes element-wise multiplication, and $s_k$ scales each interaction term. By default, these scales are learned jointly with the policy.
\begin{itemize}[leftmargin=*,itemsep=1pt,topsep=1pt]
\item{$k=1$:} $\psi_1$ is an affine feature, as in a standard MLP.
\item{$k=2$:} $\psi_2$ retains the affine path while adding products of two independently learned affine features, $A_1$ and $A_2$. Tying their weights and biases would restrict each product to the square of a single affine projection. Separate weights allow interactions between different learned input directions, capturing couplings across joints, observation channels, and history frames. These interactions are learned from the task objective rather than assigned predefined physical meanings.
\item{$k\ge 2$:} In general, each step multiplies the preceding feature by a new affine feature and adds the result, extending the polynomial degree without enumerating the full monomial basis.
\end{itemize}
Setting scales $s_2,\ldots,s_K$ to zero gives $\psi_K\!=\!A_1$, reducing \sname{} to a standard MLP. Thus, \sname{} replaces only the first affine layer with $\psi_K$, leaving subsequent nonlinear operations $g$ unchanged:
\begin{equation}
z_t = g\!\left(\psi_K(x_t)\right).
\label{eq:prism_mapping}
\end{equation}
Multiplication precedes these nonlinear operations: $\psi_K$ is polynomial in $x_t$, whereas $z_t$ need not be. The policy combines $z_t$ with existing features derived from $c_t$, when present.

\paragraph{\sname{} for reinforcement learning.}
\label{sec:method_rl}

For reinforcement learning, \sname{} replaces the
state or history projection feeding the actor.
The polynomial encoder constructs $\psi_K(x_t)$,
and $g$ maps it to the feature dimension
expected by the remaining actor network.
The action head retains its original
parameterization, and predicted actions are executed
through the existing low-level controller.

The polynomial encoder and actor are optimized jointly
using the policy's native reinforcement-learning
objective.
The observation history, reward definition, and
action interface are unchanged.
When privileged observations are used for critic
training, they remain restricted to value estimation
and are not provided to the polynomial encoder
or deployed actor.

\paragraph{\sname{} for imitation learning.}
For imitation learning, \sname{} replaces the
proprioceptive conditioning projection while preserving
its output dimension and downstream interface.
Given the proprioceptive input or history $x_t$,
we compute
$z_t^{\mathrm{proprio}}=g(\psi_K(x_t))$.
This representation is combined with the existing
visual and task-conditioning features through the
policy's original conditioning mechanism. The polynomial representation is optimized jointly
with the policy's trainable components using its
native imitation-learning objective.
The visual and language pathways, action-generation
process, and low-level controller are retained. Implementation-specific input configurations, projection networks, and training settings are provided in the Appendix.
\def\tabLoco#1{
\begin{table}[#1]
\centering
\caption{
\textbf{\sname{} improves humanoid locomotion beyond the
tested increase in model capacity.}
Performance comparison on Humanoid-Gym~\cite{gu2024humanoid}.
Results are mean $\pm$ standard deviation over five training seeds.
The larger MLP matches the reported parameter count of \sname{}. Survival rate denotes the percentage of episodes reaching 24 seconds. Bold indicates the best reported mean.
At the same parameter count as MLP-L,
\sname{} increases survival from 52.25\% to 92.50\%,
showing that the improvement is not explained
by increased model capacity alone.}
\label{tab:locomotion_main}
\resizebox{\columnwidth}{!}{%
\begin{tabular}{lcccccc}
\toprule
Method
& Interaction
& \#Parameters
& Episode Length $\uparrow$
& Linear Error $\downarrow$
& Yaw Error $\downarrow$
& Survival Rate $\uparrow$ \\
\midrule
MLP
& \xmark
& 0.926M
& $1340.8 \pm 39$
& $0.4607 \pm 0.011$
& $0.2817 \pm 0.008$
& $51.00 \pm 3.7$ \\
MLP-L
& \xmark
& 1.321M
& $1349.9 \pm 29$
& $0.4738 \pm 0.028$
& $0.3086 \pm 0.010$
& $52.25 \pm 2.9$ \\
\midrule
\sname{}
& \cmark
& 1.321M
& $\mathbf{2233.4 \pm 23}$
& $\mathbf{0.2099 \pm 0.008}$
& $\mathbf{0.1221 \pm 0.005}$
& $\mathbf{92.50 \pm 2.4}$ \\
\bottomrule
\end{tabular}}
\end{table}
}

\def\tabMani#1{
\begin{table}[#1]
\centering
\caption{
\textbf{\sname{} achieves higher success rates than ACT
across all three manipulation tasks.}
We compare ACT~\cite{zhao2023learning}
and \sname{} on three ManiSkill3
tasks~\cite{tao2025maniskill3}, using CroSTAta recovery
demonstrations~\cite{minelli2025crostata}.
Success rates (\%) use the final 100K checkpoint,
one training seed per task, and five evaluation seeds.
Average is the mean success rate across the three tasks. \sname{} achieves higher success on all three tasks,
increasing the average from 63.5\% to 67.6\%.
These gains under matched training conditions
support the benefit of the representation
across different manipulation tasks.
}
\label{tab:dp_manipulation_success}
\small
\setlength{\tabcolsep}{17pt}
\begin{tabular}{@{\hspace{6pt}}lcccc@{\hspace{6pt}}}
\toprule
Method & StackCube & TwoRobotStackCube & G1TransportBox & Average \\
\midrule
ACT~\cite{zhao2023learning}
& $58.8 \pm 6.3$
& $47.0 \pm 3.2$
& $84.8 \pm 4.9$
& $63.5$ \\
\sname{}
& $\mathbf{68.8 \pm 5.8}$
& $\mathbf{48.2 \pm 2.8}$
& $\mathbf{85.8 \pm 5.1}$
& $\mathbf{67.6}$ \\
\bottomrule
\end{tabular}
\end{table}
}

\def\tabTrackMultiTask#1{
\begin{table}[#1]
\centering
\begin{minipage}[t]{0.44\linewidth}
\vspace{0pt}
\centering
\caption{
\textbf{\sname{} improves humanoid reference-motion tracking.}
We compare BFM-Zero~\cite{li2025bfm} and its variants
at aligned checkpoints after 9.6M environment steps,
using one training seed.
Nominal uses dynamic friction
coefficients of 1.0 and the original robot masses.
Params are in millions; bold marks the lowest
tracking EMD.
}
\label{tab:bfm_zero_capacity}
\small
\setlength{\tabcolsep}{7pt}
\renewcommand{\arraystretch}{1.08}
\begin{tabular}{lcc}
\toprule
Method & \#Parameters & Nominal EMD $\downarrow$ \\
\midrule
MLP & 32M & 1.104 \\
\rowcolor{gray!10}
MLP-L & 49M & 1.090 \\
\midrule
\textbf{\sname{}} & 32 M& \textbf{1.050} \\
\bottomrule
\end{tabular}
\end{minipage}\hfill
\begin{minipage}[t]{0.54\linewidth}
\vspace{0pt}
\centering
\caption{
\textbf{\sname{} improves mean success in multi-task manipulation.}
We compare SmolVLA~\cite{shukor2025smolvla} and its variants
on LIBERO at aligned 80K-update checkpoints.
Success (\%) is mean $\pm$ SD over three training seeds.
The larger control replaces the linear state projection
with a larger MLP.
Params are in millions; $\Delta$ denotes percentage-point
gains over Baseline. Bold marks the highest mean. 
}
\label{tab:smolvla_results}
\small
\setlength{\tabcolsep}{3pt}
\renewcommand{\arraystretch}{1.08}
\begin{tabular}{lccc}
\toprule
Method
& \#Parameters
& LIBERO Average $\uparrow$
& $\Delta$ \\
\midrule
SmolVLA
& 450M & $62.00 \pm 1.67$ & -- \\
\rowcolor{gray!10}
Larger Model
& 456M & $62.57 \pm 3.14$ & $+0.57$ \\
\midrule
\textbf{\sname{}}
& 451M
& $\mathbf{63.82 \pm 2.59}$
& \textbf{+1.82} \\
\bottomrule
\end{tabular}
\end{minipage}
\end{table}
}

\vspace{-0.5em}
\section{Experimental Results}
\label{sec:results}
\vspace{-0.5em}

We ask four questions:
{\bf 1)} Does \sname{} improve locomotion and manipulation performance over standard and larger baselines?
{\bf 2)} Does \sname{} improve contact behavior and real-robot manipulation without additional sensing or controller changes?
{\bf 3)} Does its advantage persist under out-of-distribution shifts in dynamics?
{\bf 4)} What physical information can be extracted from its learned representations, and how do multiplicative features affect action prediction?

\subsection{Task Performance and Capacity Controls}
\label{sec:locomotion_results}

We evaluate whether \sname{} improves policy performance
across locomotion and manipulation backbones, and whether
increasing baseline capacity reproduces these gains.
Within each backbone comparison, methods share the
observation and action interfaces, training data or
environment, native learning objective, and low-level
controller.

For locomotion, we evaluate velocity-command tracking
on flat terrain in Humanoid-Gym~\cite{gu2024humanoid}
using MLP actors trained with PPO~\cite{schulman2017proximal}.
We report episode length, linear- and yaw-velocity
tracking errors, and 24\,s survival rate.
We also evaluate BFM-Zero~\cite{li2025bfm} on
reference-motion tracking using Earth Mover's Distance (EMD).
For manipulation, we evaluate task-specific
ACT~\cite{zhao2023learning} on three tasks in ManiSkill3~\cite{tao2025maniskill3}, a physics-based
manipulation benchmark, and multi-task
SmolVLA~\cite{shukor2025smolvla} on
LIBERO~\cite{liu2023libero}, a language-conditioned
manipulation benchmark, reporting task success.

\tabLoco{!b}
\tabMani{!b}

\tabTrackMultiTask{!t}

\paragraph{Performance across diverse policy backbones.}
In Humanoid-Gym and ManiSkill3 manipulation,
\sname{} improves the reported locomotion metrics
and task success across all evaluated tasks,
respectively
(Tables~\ref{tab:locomotion_main}
and~\ref{tab:dp_manipulation_success}).
The benefit also extends to reference-motion
tracking~(Table~\ref{tab:bfm_zero_capacity}) and multi-task SmolVLA~(Table~\ref{tab:smolvla_results}), with lower tracking
EMD and higher mean LIBERO success.

{\bf Comparison with increased baseline capacity.}
For Humanoid-Gym, BFM-Zero, and SmolVLA, we next
test whether increasing baseline capacity reproduces
these gains.
The controls widen the Humanoid-Gym actor and
BFM-Zero main network, and replace SmolVLA's linear
state projection with a larger MLP.
The Humanoid-Gym control matches the reported
parameter count of \sname{}, whereas the BFM-Zero
and SmolVLA controls use more parameters.
Nevertheless, these controls do not match the
reported performance of \sname{}
(Tables~\ref{tab:locomotion_main},
\ref{tab:bfm_zero_capacity},
and~\ref{tab:smolvla_results}).

\subsection{Practical Benefits for Robot Control}
\label{sec:contact_results}

We examine the execution behavior of \sname{} in
contact-rich simulation and its practical benefits
on a physical SO-101~\cite{knight2025standard} robot.

\begin{figure*}[b]
\centering
\setlength{\tabcolsep}{0pt}
\begin{tabular*}{\linewidth}{@{\extracolsep{\fill}}ccccc@{}}
\footnotesize{Contact} &
\footnotesize{ACT} &
\footnotesize{MCC-Sensorless} &
\footnotesize{MCC-Oracle} &
\footnotesize{\sname{}} \\[2pt]
\includegraphics[height=0.160\linewidth]{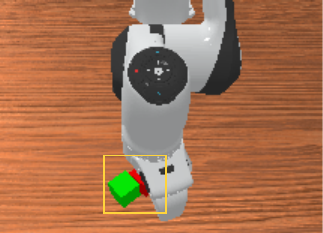} &
\includegraphics[height=0.160\linewidth]{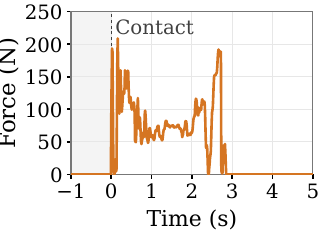} &
\includegraphics[height=0.160\linewidth]{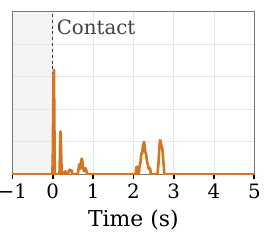} &
\includegraphics[height=0.160\linewidth]{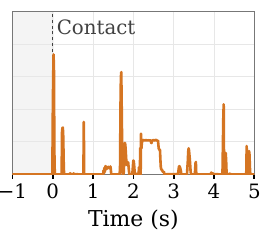} &
\includegraphics[height=0.160\linewidth]{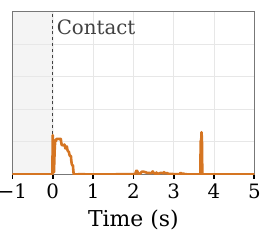}
\end{tabular*}%

\vspace{-0.4em}
\caption{
\textbf{\sname{} exhibits compliance-like contact behavior
in the illustrated recovery.}
The leftmost image illustrates hand--cube contact
during recovery from an induced object drop
in StackCube~\cite{tao2025maniskill3}.
The remaining panels compare hand--cube contact force
($F$, orange) for ACT~\cite{zhao2023learning}, MCC-Sensorless~\cite{shi2026minimalist}, MCC-Oracle~\cite{shi2026minimalist},
and \sname{}, respectively, from the same initial state.
All force traces are aligned to first contact at $t=0$
and use identical axes.
In this example, \sname{} rapidly relaxes contact force
and first completes recovery 0.85 seconds after contact;
the low-load phase includes object release.
Execution continues for five seconds after contact,
including after task completion.
Contact force is recorded only for evaluation
and is not provided to the \sname{} actor.
}
\label{fig:force_compliance}
\end{figure*}

\paragraph{Policy representation versus execution-time correction.}
We investigate whether execution-time contact corrections
applied to a fixed baseline policy can achieve
similar task performance to \sname{}.
We compare against two simulation adaptations of
Minimalist Compliance Control (MCC)~\cite{shi2026minimalist}.
MCC-Sensorless applies force-proxy corrections to
the translation commands of a frozen ACT policy,
whereas MCC-Oracle uses simulator contact forces.
{\bf Fig.~\ref{fig:force_compliance} compares contact-force}
traces from a representative StackCube rollout in ManiSkill3,
aligned to the first hand--cube contact.
In this example, \sname{} rapidly reduces contact
force following contact and subsequently completes
the task.
Importantly, \sname{} receives no force, wrench,
or tactile feedback, and uses no explicit
admittance controller.
{\bf Table~\ref{tab:libero_mcc_comparison} compares induced-drop
recovery} on 44 shared validated failure states.
\sname{} achieves the highest observed recovery rate (52.3\%),
compared with 40.9\% for ACT, 13.6\% for MCC-Sensorless,
and 34.1\% for MCC-Oracle, which uses simulator-force feedback.
\sname{} also has the lowest mean capped recovery time
(3.11 seconds) and terminal placement error (96.2 mm).
These summaries include all 44 attempts; unrecovered cases
are assigned the 5 seconds horizon for recovery time.

\begin{table*}[!t]
\centering

\begin{minipage}[t]{0.48\textwidth}
\caption{
\textbf{\sname{} improves failure recovery without
explicit admittance control.}
}
\label{tab:libero_mcc_comparison}

\vspace{0.3em}
\centering
\footnotesize
\setlength{\tabcolsep}{2.5pt}
\renewcommand{\arraystretch}{1.15}
\begin{tabular*}{\linewidth}{lccc}
\toprule
Method & Recovery$\uparrow$ & Time(s)$\downarrow$
& Error(mm)$\downarrow$\\
\midrule
ACT~\cite{zhao2023learning}
& 40.9\% & 3.38 & 141.7 \\
+MCC-Sensorless
& 13.6\% & 4.50 & 123.5 \\
+MCC-Oracle
& 34.1\% & 3.71 & 105.4 \\
\midrule
\sname{}
& \textbf{52.3\%} & \textbf{3.11} & \textbf{96.2} \\
\bottomrule
\end{tabular*}

\par\vspace{0.4em}
\begin{minipage}{\linewidth}
\normalsize
\textit{Note.}
All methods use the same 44 validated drop states
and final 100K checkpoints.
Recovery is task completion rate.
Time and placement error include failures;
failed trials receive a capped recovery time of 5\,s.
MCC variants~\cite{shi2026minimalist} are simulation adaptations.
\end{minipage}
\end{minipage}
\hfill
\begin{minipage}[t]{0.48\textwidth}
\caption{
\textbf{\sname{} improves real-robot SO-101 tape-pickup success.}
}
\label{tab:so101_tape}

\vspace{0.3em}
\centering
\footnotesize
\setlength{\tabcolsep}{5pt}
\renewcommand{\arraystretch}{1.15}
\begin{tabular*}{\linewidth}{lccc}
\toprule
Method & Success$\uparrow$ & Motion($^\circ$)$\uparrow$& Error($^\circ$)$\downarrow$ \\
\midrule
ACT~\cite{zhao2023learning}
& 68.0\% & $31.1 \pm 2.9$ & $27.4 \pm 2.9$ \\
\sname{}
& \textbf{94.0\%} & $\mathbf{38.5 \pm 1.5}$
& $\mathbf{24.3 \pm 4.2}$ \\
\bottomrule
\end{tabular*}

\par\vspace{0.4em}
\begin{minipage}{\linewidth}
\normalsize
\textit{Note.}
Both policies use the same 50 demonstrations
and are evaluated over 50 hardware rollouts each.
Joint motion measures movement during the first 3 seconds;
larger values indicate more initial movement.
Tracking error measures the difference between
commanded and measured joint angles;
lower values indicate closer tracking.
\end{minipage}
\end{minipage}

\end{table*}

\paragraph{SO-101 tape pickup.}
To assess benefits on hardware, we compare ACT~\cite{zhao2023learning} and
\sname{} on a physical SO-101 robot using the
same demonstrations and hardware protocol.
The task requires lifting the tape roll at least
5\,cm above the table and maintaining that height
for 3 seconds.
We report task success, arm joint-space travel during
the first 3 seconds, and rollout-averaged command--state error.
Table~\ref{tab:so101_tape} shows higher pickup success
for \sname{}, accompanied by greater early
joint-space travel and lower mean command--state error.

\subsection{Generalizing to Out-of-distribution Physical Conditions}
\label{sec:robustness_results}

We test whether \sname{} retains its performance advantage
under changes in physical dynamics.
All evaluations use fixed trained checkpoints without retraining.
We report 60-second survival on flat terrain in Humanoid-Gym,
20-second supported survival on random rough terrain in
mjlab~\cite{zakka2026mjlab}, and reference-motion tracking
Earth Mover's Distance (EMD) for BFM-Zero.
Nominal dynamics provide the reference condition
within each evaluation protocol.

\begin{figure*}[!t]
\centering
\begingroup
\setlength{\tabcolsep}{0pt}
\begin{tabular}{@{}cccc@{}}
\multicolumn{4}{@{}l@{}}{\footnotesize\textbf{Humanoid-Gym (flat terrain)}} \\[-2pt]
\includegraphics[width=0.25\linewidth]{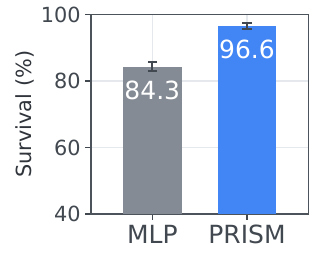} &
\includegraphics[width=0.25\linewidth]{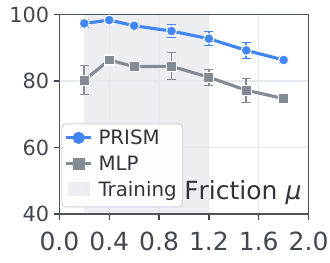} &
\includegraphics[width=0.25\linewidth]{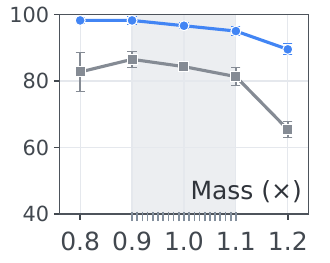} &
\includegraphics[width=0.25\linewidth]{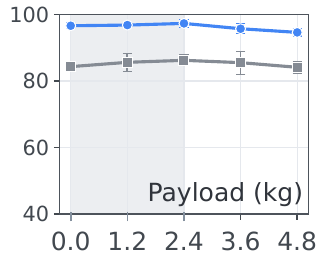} \\
[-3pt]
\multicolumn{4}{@{}l@{}}{\footnotesize\textbf{mjlab (rough terrain)}} \\[-2pt]
\includegraphics[width=0.25\linewidth]{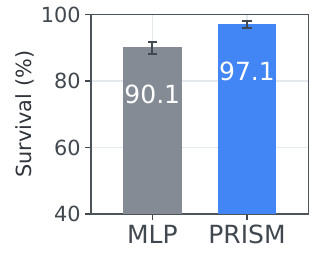} &
\includegraphics[width=0.25\linewidth]{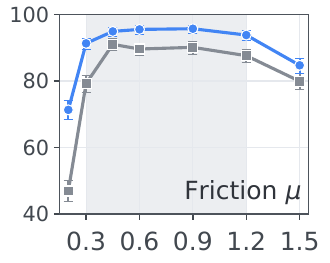} &
\includegraphics[width=0.25\linewidth]{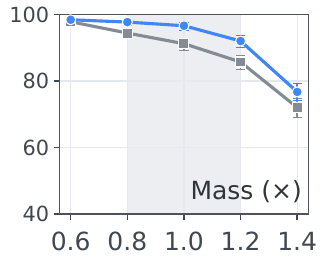} &
\includegraphics[width=0.25\linewidth]{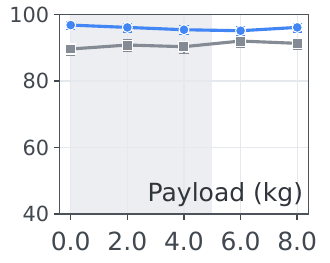} \\
\end{tabular}
\endgroup
\vspace{-1em}
\caption{\textbf{Humanoid survival under dynamics changes.}
Columns show survival under nominal dynamics and changes in friction,
robot mass, and payload.
Rows show flat terrain in Humanoid-Gym~\cite{gu2024humanoid} (top) and random rough terrain
in mjlab~\cite{zakka2026mjlab} (bottom).
Shading marks training ranges or discrete-value envelopes. \sname{} achieves higher survival than the MLP in all displayed conditions.}
\label{fig:locomotion_dynamics}
\label{fig:hg_robustness_curves}
\label{fig:mjlab_rough_dynamics}
\end{figure*}

\paragraph{Locomotion under dynamics changes.}
We compare \sname{} with a parameter-matched MLP
in Humanoid-Gym and with a standard MLP in mjlab.
We vary friction, robot-link mass/inertia, and payload
one factor at a time, including values outside
their respective training ranges.
\sname{} achieves higher survival than a parameter-matched MLP
on flat terrain in Humanoid-Gym and a standard MLP on random
rough terrain in mjlab~\cite{zakka2026mjlab} across all displayed
conditions (Fig.~\ref{fig:locomotion_dynamics}).

Across six settings beyond each simulator's training ranges, \sname{} achieves mean survival of 92.3\% versus 78.3\% for a parameter-matched MLP on flat terrain in Humanoid-Gym, and 87.1\% versus 80.0\% for a standard MLP on random rough terrain in mjlab. In Humanoid-Gym, these settings span friction coefficients of 1.5--1.8, mass scaling factors of 0.8--1.2, and payloads of 3.6--4.8\,kg. The corresponding ranges in mjlab are 0.2--1.5, 0.6--1.4, and 6--8\,kg.

We evaluate the BFM-Zero policies under low friction and increased link mass, using the same checkpoints
as in the nominal evaluation.
Table~\ref{tab:bfm_zero_robustness} shows that \sname{}
retains lower tracking EMD than both the original
and larger-capacity BFM-Zero backbones
under both dynamics changes.

\begin{table}[!t]
\centering
\caption{
\textbf{\sname{} retains lower reference-motion tracking
error under dynamics changes.}
We report tracking EMD for BFM-Zero~\cite{li2025bfm}
and its variants at fixed checkpoints after
9.6M environment steps, using one training seed.
Nominal uses terrain static and dynamic friction
coefficients of 1.0 and the original link masses.
Low-friction sets both coefficients to 0.20;
Robot Mass scales link masses by 1.15.
Each perturbation changes one factor from nominal.
Mean averages the three conditions;
bold marks the lowest EMD in each column. Both perturbations increase EMD for every model,
but \sname{} remains the lowest in each condition,
showing that its tracking advantage persists
beyond nominal dynamics.
}
\label{tab:bfm_zero_robustness}
\small
\setlength{\tabcolsep}{18pt}
\renewcommand{\arraystretch}{1.08}
\begin{tabular}{lcccc}
\toprule
Method
& Nominal $\downarrow$
& Low-friction $\downarrow$
& Robot Mass $\downarrow$
& Mean $\downarrow$ \\
\midrule
Baseline & 1.104 & 1.582 & 1.121 & 1.269 \\
\rowcolor{gray!10}
Larger Model & 1.090 & 1.589 & 1.114 & 1.264 \\
\midrule
\textbf{\sname{}}
& \textbf{1.050}
& \textbf{1.548}
& \textbf{1.073}
& \textbf{1.224} \\
\bottomrule
\end{tabular}
\vspace{-0.5em}
\end{table}

\paragraph{Robustness to manipulation failures.}
We evaluate whether \sname{} recovers from
induced object drops in ManiSkill3
StackCube~\cite{tao2025maniskill3}.
Using ACT~\cite{zhao2023learning} and its
\sname{} variant trained on the same CroSTAta
recovery demonstrations~\cite{minelli2025crostata},
we measure regrasp-and-lift and subsequent
task completion within 100 control steps.
\sname{} achieves higher regrasp (75.0\% vs. 68.2\%)
and recovery success (52.3\% vs. 40.9\%)
than ACT.
Further details are provided in
Appendix~\ref{sec:appendix_execution_perturbations}.

\subsection{Learned Interactions and Action Prediction}

We examine the information accessible from learned representations, the influence of multiplicative features on action prediction, and the impact of polynomial degree on policy performance.

\paragraph{Physical-response probing.}
We fit linear probes on frozen policy representations
and evaluate predictions on held-out rollout windows
with horizon $H=5$.
The targets include slip, joint power, contact impulse,
and a contact-work proxy, computed from simulator
states, contacts, and actions only for analysis.
We report mean squared error (MSE) and Pearson
correlation coefficient (PCC).
Table~\ref{tab:physics_probe_by_domain} shows lower
MSE and higher PCC for \sname{} across the evaluated
response proxies.
The contact probes complement the contact-behavior
analysis in Section~\ref{sec:contact_results}
by assessing the linear accessibility of
force-related response information.

\paragraph{Action sensitivity to multiplicative features.}
Linear predictability alone does not establish feature use
in action prediction.
We therefore remove individual latent products from the
frozen policies while retaining the observations, affine
path, learned weights, and other products.
Table~\ref{tab:prism_learned_terms} highlights two structures:
temporal and cross-joint factor dependence.
In humanoid tracking, the leading factors emphasize
left-hip velocity across consecutive observations and
left--right hip velocities; removing their products
changes predicted actions by 0.057 and 0.025, respectively.
In TwoRobotStackCube, the leading factors span the two
arms and an arm--gripper history, with corresponding
action changes of 0.00338 and 0.00270.
These interventions show that learned multiplicative
features influence action generation in both domains.
The displayed input products summarize dense factors,
rather than identified physical mechanisms.


\paragraph{Polynomial degree.}
We compare affine, quadratic, and cubic representations
under the same Humanoid-Gym setup
(Appendix Table~\ref{tab:degree_ablation_humanoid}).
Quadratic interactions capture most of the observed
performance gain over the affine variant, while
the cubic variant improves further.
We use the quadratic representation as the default
for its simpler construction.

\begin{table}[!t]
\centering
\caption{
\textbf{\sname{} representations make mechanics-inspired
responses more linearly predictable.}
We conduct linear probing by freezing trained policy
representations, fitting linear readouts, and evaluating
them on held-out windows with horizon $H=5$.
Targets use simulator measurements only for analysis.
Here, $t$ is the window start and $h$ indexes future samples;
$\mathcal F$ contains the two feet,
$c_{f,t}$ is one when foot $f$ has vertical contact force
above 5\,N and zero otherwise, and
$v^{\mathrm{world}}_{f,xy,t}$ is its world-frame planar velocity.
The symbols $\tau_{j,t}$ and $\dot q_{j,t}$ denote
joint-$j$ torque and angular velocity.
For the contact targets, $F_t$ is the scalar contact-load
signal and $\Delta x_t$ the spatial-displacement vector
used to construct the proxy; norms are Euclidean
and $\log$ denotes the natural logarithm.
The contact rows retain the reported discrete proxy formulas,
not calibrated continuous-time impulse or work estimates.
MSE is mean squared error; PCC is the Pearson correlation coefficient. Lower MSE and higher PCC across all four targets
indicate that mechanics-inspired response information
is more linearly accessible in \sname{} representations.
}
\label{tab:physics_probe_by_domain}
\renewcommand{\arraystretch}{1.25}
\setlength{\tabcolsep}{3.5pt}
\definecolor{WinGreen}{HTML}{E2F0D9}

\resizebox{\columnwidth}{!}{%
\begin{tabular}{llcccccc}
\toprule
&
& \multicolumn{2}{c}{MSE $\downarrow$}
& \multicolumn{2}{c}{PCC $\uparrow$}
& \multicolumn{2}{c}{Relative Gain} \\
\cmidrule(lr){3-4}
\cmidrule(lr){5-6}
\cmidrule(lr){7-8}
Target & Formula
& MLP & \sname{}
& MLP & \sname{}
& MSE & PCC \\
\midrule
Slip velocity
& $\sum_{h=1}^{H}\sum_{f\in\mathcal F}
c_{f,t+h}\|v^{\mathrm{world}}_{f,xy,t+h}\|_2$
& 0.711 & \textbf{0.700}
& 0.501 & \textbf{0.549}
& \cellcolor{WinGreen}\textbf{$-$1.5\%}
& \cellcolor{WinGreen}\textbf{$+$9.6\%} \\

Joint power
& $\log\!\left(
1+\sum_{h=1}^{H}\sum_{j=1}^{12}
|\tau_{j,t+h}\dot q_{j,t+h}|
\right)$
& 0.328 & \textbf{0.282}
& 0.820 & \textbf{0.848}
& \cellcolor{WinGreen}\textbf{$-$14.0\%}
& \cellcolor{WinGreen}\textbf{$+$3.4\%} \\

\midrule
Contact impulse
& $\log(1+\sum_{h\le H}F_{t+h})$
& 0.301 & \textbf{0.241}
& 0.829 & \textbf{0.861}
& \cellcolor{WinGreen}\textbf{$-$19.9\%}
& \cellcolor{WinGreen}\textbf{$+$3.9\%} \\

Contact work
& $\sum_{h\le H}F_{t+h}\|\Delta x_{t+h}\|$
& 0.362 & \textbf{0.319}
& 0.755 & \textbf{0.780}
& \cellcolor{WinGreen}\textbf{$-$11.9\%}
& \cellcolor{WinGreen}\textbf{$+$3.3\%} \\
\bottomrule
\end{tabular}}
\vspace{-0.2em}
\end{table}

\begin{table}[t!]
\centering
\caption{
\textbf{\sname{} uses temporal and cross-joint features
in action prediction.}
We measure action changes after removing one latent product.
For Humanoid-Gym, $\Delta a$ is the mean $L_2$ distance
between joint-target vectors after native action scaling,
reported in radians.
For ACT, it is the coordinate-wise mean absolute change
in the first predicted normalized action.
The two statistics have different units and are not
directly comparable across domains.
Products summarize dominant inputs to dense affine factors. With the observations and affine paths unchanged,
removing these products alters actions in both tasks,
showing that the learned multiplicative features
contribute to action prediction.
}
\label{tab:prism_learned_terms}
\label{tab:interaction_action_sensitivity}
\small
\setlength{\tabcolsep}{4pt}
\renewcommand{\arraystretch}{1.12}
\begin{tabular*}{\linewidth}{@{\extracolsep{\fill}}lllc@{}}
\toprule
Task Context & Top-ranked Interaction & Interaction Type & $\Delta a$ \\
\midrule
Humanoid velocity tracking
& $\dot q_{\mathrm{Lhippitch}}(t)
   \times\dot q_{\mathrm{Lhippitch}}(t-1)$
& Velocity--history & 0.057 \\
& $\dot q_{\mathrm{Rhippitch}}(t)
   \times\dot q_{\mathrm{Lhippitch}}(t)$
& Cross-joint velocity & 0.025 \\
\midrule
TwoRobotStackCube
& $q_{\mathrm{L5}}(t)\times q_{\mathrm{R7}}(t)$
& Cross-arm position & 0.00338 \\
& $q_{\mathrm{L5}}(t-1)\times g_{\mathrm{R2}}(t)$
& Arm--gripper history & 0.00270 \\
\bottomrule
\end{tabular*}
\vspace{-0.5em}
\end{table}



\vspace{-1em}
\section{Conclusion}
\vspace{-1em}

\sname{} shows that better robot control is not simply a matter of larger networks. Its compact polynomial layer learns multiplicative interactions among existing observations without prescribing robot dynamics or changing training objectives, sensing, or controllers. Across locomotion and manipulation, it improves performance, including on hardware, and retains its advantage under changes in friction, mass, and payload beyond the training ranges. These results show how changing the representation of observations can improve robot policies without redesigning the control system.

\noindent\textbf{Limitations.}  \sname{} relies on available observations to learn useful interactions; its benefits may be limited when essential information is missing.

\clearpage
\vspace{-1em}

\section*{Acknowledgements}
\vspace{-1em}
This project was supported in part by NSF grant 2313151, with additional computing resources provided through NAIRR Pilot CIS240421 and CIS240431.
We thank WHATs LAB in South Korea for generously lending robots that enabled our real-world manipulation experiments there.

\bibliography{example.bib}

\clearpage

\appendix

\section{Task Performance and Capacity Controls}
\label{sec:appendix_rq1}
\label{sec:appendix_experimental_details}
\label{sec:appendix_more_results}

This appendix follows the experimental organization of the main text:
task performance and capacity controls, practical robot control,
generalization to changed physical conditions, and representation analysis.
We use the main-text partition $o_t=(x_t,c_t)$, where $x_t$ is the
observation stream processed by \sname{} and $c_t$ contains the remaining
inputs. Each affine factor $A_k(x_t)=W_kx_t+b_k$ maps
$\mathbb R^d$ to a common latent space $\mathbb R^m$, with
$W_k\in\mathbb R^{m\times d}$ and $b_k\in\mathbb R^m$.
The symbol $g$ denotes the feature mapping between the polynomial block
and the remaining policy, as illustrated in the main architecture figure.
Its projection, normalization, and activation operations are specified
for each backbone below. Scalar scales are broadcast over latent
coordinates; channel-wise scales are written explicitly with $\odot$.
The polynomial degree describes the feature before nonlinear operations,
not the complete policy.

\subsection{Humanoid-Gym locomotion}
\label{sec:appendix_humanoid_gym}

For the flat-terrain Unitree G1 comparison in
Humanoid-Gym~\cite{gu2024humanoid} (Table~\ref{tab:locomotion_main}),
Tables~\ref{tab:appendix_locomotion_details},
\ref{tab:appendix_locomotion_reward}, and
\ref{tab:appendix_locomotion_capacity} report the training
settings, reward weights, and actor parameter counts.
The actor input $x_t\in\mathbb{R}^{705}$ concatenates 15 frames
of 47-dimensional proprioceptive and command observations.

\paragraph{Polynomial encoder and actor.}
Humanoid-Gym uses a residual parameterization of the affine-plus-product
representation. To connect its implementation to the affine factors in
the main text while retaining its actual projections, define
\begin{align}
 \ell_t&=\mathcal H(x_t), & p_{0,t}&=\mathcal B_0(\ell_t),
 \label{eq:appendix_hg_affine}\\
 A_1^{\mathrm{HG}}(x_t)&=\mathcal B_1(\ell_t), &
 A_2^{\mathrm{HG}}(x_t)&=\mathcal S_0(p_{0,t}),\nonumber\\
 r_t&=A_1^{\mathrm{HG}}(x_t)\odot A_2^{\mathrm{HG}}(x_t), &
 \psi_2^{\mathrm{HG}}(x_t)&=p_{0,t}+r_t,
 \label{eq:appendix_hg_product}\\
 z_t&=\operatorname{ELU}\!\left(
 \mathcal R(x_t)+\beta_{\mathrm{poly}}
 \mathcal P(\psi_2^{\mathrm{HG}}(x_t))\right).
 \label{eq:appendix_hg_encoder}
\end{align}
All six calligraphic maps are learned affine maps, including biases.
The maps $\mathcal H$ and $\mathcal R$ take 705 inputs to 256 outputs;
$\mathcal B_0$, $\mathcal B_1$, $\mathcal S_0$, and $\mathcal P$
each map 256 features to 256 features. Thus both
$A_1^{\mathrm{HG}}$ and $A_2^{\mathrm{HG}}$ are affine in $x_t$,
and $r_t$ contains products of learned affine observations.
The superscript distinguishes this implementation from the compact
construction: its affine base is $p_{0,t}$, and it additionally retains
$\mathcal R(x_t)$. These paths and projections are included in the
reported parameter counts, rather than identified with a bare pair of
$705\!\to\!256$ factors.

The scalar $\beta_{\mathrm{poly}}$ increases from zero to one over
500 proximal policy optimization (PPO)~\cite{schulman2017proximal}
updates and scales the polynomial branch relative to the raw-input
affine path. It is a scheduled buffer, not a learned interaction scale
or the generalized advantage estimation
(GAE)~\cite{schulman2015high} parameter $\lambda$.
At zero scale, the encoder becomes
$z_t=\operatorname{ELU}(\mathcal R(x_t))$.
The 256-dimensional output feeds the actor MLP
$256\to512\to256\to128\to12$, with exponential linear unit
(ELU) activations~\cite{clevert2016fast} after each hidden layer
and no input or hidden LayerNorm. Its final projection predicts the mean
of a Gaussian distribution over joint-position targets. The native
action interface and low-level proportional--derivative (PD) controller
are retained.

\begin{table}[!htbp]
\centering
\caption{Humanoid-Gym~\cite{gu2024humanoid} training configuration for
Table~\ref{tab:locomotion_main}.
Ten physics steps per control tick give a control rate of 100\,Hz.}
\label{tab:appendix_locomotion_details}
\footnotesize
\setlength{\tabcolsep}{3pt}
\renewcommand{\arraystretch}{1.05}
\begin{tabular}{@{}ll@{}}
\toprule
Item & Setting \\
\midrule
Actor observation & $15{\times}47$ proprioceptive/command history \\
Critic observation & $3{\times}73$ privileged history \\
Action & 12-dimensional joint-position target, scale $0.25$ \\
Simulation & $\Delta t=0.001\,\mathrm{s}$, control decimation $10$ \\
PD gains & $K_p=\{100,150,40\}$, $K_d=\{2,4,2\}$ \\
Linear commands (m/s) & $v_x\!\in[-0.3,0.6]$, $v_y\!\in[-0.3,0.3]$ \\
Yaw-rate command (rad/s) & $\omega_z\!\in[-0.3,0.3]$ \\
Randomization & Friction, mass, pushes, delay, action noise \\
Learning algorithm & PPO~\cite{schulman2017proximal}, learning rate $10^{-5}$ \\
Discount $\gamma$ / GAE $\lambda$ & $0.994$ / $0.9$ \\
Rollout / epochs / minibatches & $60$ / $2$ / $4$ \\
Training iterations & $3001$ \\
\sname{} degree / latent width & $2$ / $256$ \\
\bottomrule
\end{tabular}
\end{table}

\paragraph{Training and performance measures.}
In Table~\ref{tab:appendix_locomotion_details}, $K_p$ and $K_d$ are
proportional and derivative gains, while $v_x$, $v_y$, and $\omega_z$
are forward, lateral, and yaw-rate commands. The gain pairs $(K_p,K_d)$
are $(100,2)$ for hip yaw, roll, and pitch, $(150,4)$ for knees, and
$(40,2)$ for ankles, symmetrically on both legs. The discount factor is
$\gamma$ and the GAE parameter is $\lambda$.
Episode length counts control steps, and survival is the percentage of
episodes reaching the 24-second horizon. At control sample $t$, the
velocity errors are
\[
 e^{\mathrm{lin}}_t=
 \|v^{\mathrm{cmd}}_{xy,t}-v^{\mathrm{base}}_{xy,t}\|_2,
 \qquad
 e^{\mathrm{yaw}}_t=
 |\omega^{\mathrm{cmd}}_{z,t}-\omega^{\mathrm{base}}_{z,t}|.
\]
Here $v_{xy}$ denotes planar velocity and $\omega_z$ denotes yaw rate;
the errors are measured in m/s and rad/s, respectively.
The evaluator averages sampled errors within episodes and then aggregates
episodes; these are not squared-error or root-mean-square-error metrics.
The mean and standard deviation in Table~\ref{tab:locomotion_main}
are across five training seeds, not individual evaluation episodes.
Table~\ref{tab:appendix_locomotion_reward} lists the nonzero native
reward weights used for this comparison.

\begin{table}[!htbp]
\centering
\caption{Active reward weights for our Humanoid-Gym~\cite{gu2024humanoid}
experiments. Zero-weight terms are omitted; functional forms follow
the released experiment configuration.}
\label{tab:appendix_locomotion_reward}
\footnotesize
\setlength{\tabcolsep}{3pt}
\renewcommand{\arraystretch}{1.03}
\begin{tabular}{@{}lrlr@{}}
\toprule
Term & Scale & Term & Scale \\
\midrule
Linear velocity & $1.2$ & Yaw velocity & $1.1$ \\
Hard tracking & $0.5$ & Low-speed regulation & $0.2$ \\
Upright orientation & $1.0$ & Base height & $0.2$ \\
Velocity consistency & $0.5$ & Base acceleration & $0.2$ \\
Joint-pose tracking & $1.6$ & Default joint pose & $0.5$ \\
Feet air time & $1.0$ & Feet clearance & $1.0$ \\
Contact number & $1.2$ & Feet distance & $0.2$ \\
Knee distance & $0.2$ & Collision & $-1.0$ \\
Contact force & $-0.01$ & Foot slip & $-0.05$ \\
Action smoothness & $-0.002$ & Torque & $-10^{-5}$ \\
Joint velocity & $-5{\times}10^{-4}$ & Joint acceleration & $-10^{-7}$ \\
\bottomrule
\end{tabular}
\end{table}

\paragraph{Capacity controls.}
The baseline MLP uses hidden widths $[512,256,128]$;
MLP-L uses $[816,352,160]$.
Table~\ref{tab:appendix_locomotion_capacity} reports actor-only
counts, separately from the main comparison's parameter totals.
The exact actor counts are 527,244 for MLP, 922,092 for
MLP-L, and 921,996 for \sname{}, including affine biases.
They follow from the complete encoder above and the corresponding
actor MLPs, without counting the critic or Gaussian standard deviations.
The main-table totals additionally include the shared critic
$219\to768\to256\to128\to1$ (398,849 parameters) and
12 learned action standard deviations.
The resulting totals are 926,105, 1,320,953, and 1,320,857,
respectively, matching the displayed 0.926M, 1.321M, and 1.321M.

\begin{table}[!htbp]
\centering
\caption{Actor parameter counts for the Humanoid-Gym~\cite{gu2024humanoid} comparison.
All actors receive the same 705-dimensional observation history.
Counts refer to the actor only and are rounded;
MLP-L and \sname{} match at the reported precision.}
\label{tab:appendix_locomotion_capacity}
\footnotesize
\setlength{\tabcolsep}{8pt}
\begin{tabular}{@{}lcc@{}}
\toprule
Method & Input dimension & Actor parameters \\
\midrule
MLP & 705 & 527K \\
MLP-L & 705 & 922K \\
\sname{} & 705 & 922K \\
\bottomrule
\end{tabular}
\end{table}

\begin{figure}[!htbp]
\centering
\begin{minipage}[t]{0.49\linewidth}
\centering
\includegraphics[width=\linewidth]{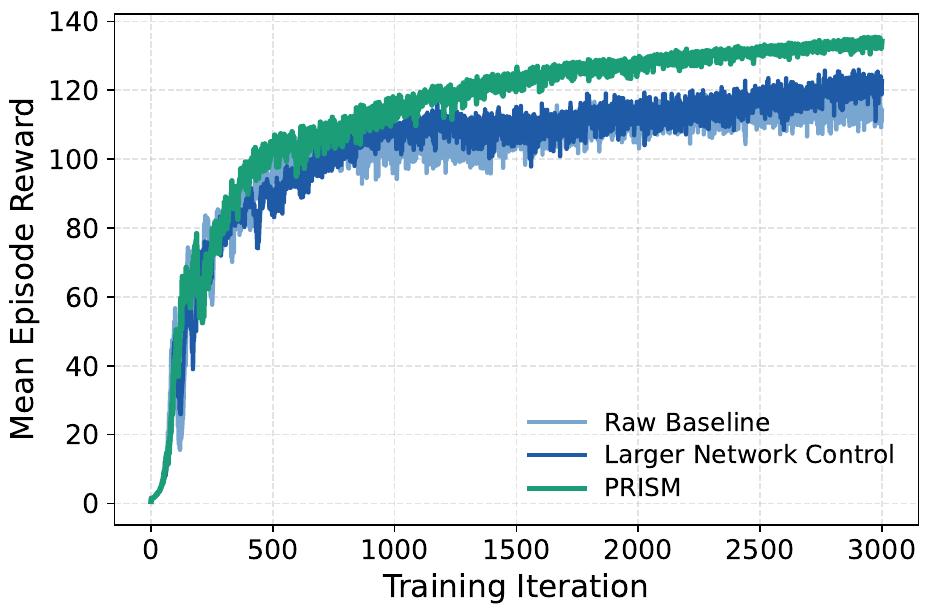}
\end{minipage}\hfill
\begin{minipage}[t]{0.49\linewidth}
\centering
\includegraphics[width=\linewidth]{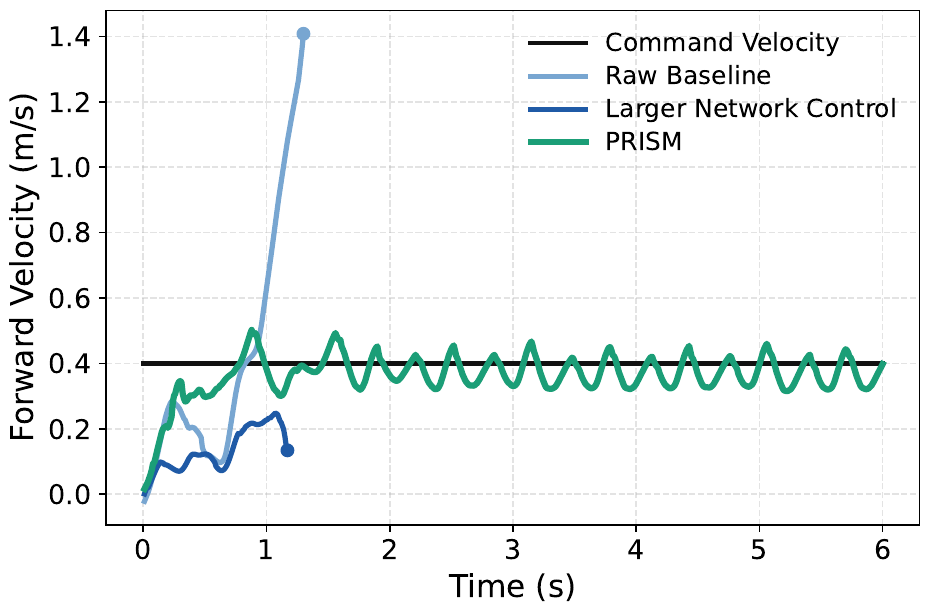}
\end{minipage}
\caption{
\textbf{\sname{} improves learning performance and illustrated command tracking.}
Left: Humanoid-Gym~\cite{gu2024humanoid} training curves with shaded
standard error across training seeds.
The legend entries Raw Baseline and Larger Network Control denote MLP
and MLP-L, respectively.
Right: realized forward velocities under a fixed command (black).
The displayed \sname{} rollout remains near the commanded velocity,
whereas the baseline traces end after falls, illustrating the
tracking and survival differences summarized in
Table~\ref{tab:locomotion_main}.
}
\label{fig:locomotion_results}
\end{figure}

Fig.~\ref{fig:locomotion_results} supplements the aggregate
results with learning curves and an illustrative rollout.

\subsection{ACT on ManiSkill3 recovery demonstrations}
\label{sec:appendix_act_maniskill}
\label{sec:appendix_diffusion_libero} 

We train ACT~\cite{zhao2023learning} and its \sname{} variant
on three ManiSkill3 tasks~\cite{tao2025maniskill3}, using the
released CroSTAta recovery demonstrations~\cite{minelli2025crostata}.
We use the demonstrations, not the CroSTAta policy model.
Each task has its own pair of policies, trained with batch size 64
and seed 1000 for 100K updates. We evaluate the final 100K
exponential-moving-average (EMA) weights, without checkpoint
selection on evaluation results.
Table~\ref{tab:appendix_act_task_contracts} specifies the data splits
and native sensor/action interfaces.

\begin{table}[!htbp]
\centering
\caption{Task-specific ACT~\cite{zhao2023learning} data and native
ManiSkill3~\cite{tao2025maniskill3} interfaces.
Train and Dev. give trajectory counts; Qpos and Action give the
measured-joint and action dimensions for one time step.
G1TransportBox uses the native \texttt{UnitreeG1TransportBox-v1} task.}
\label{tab:appendix_act_task_contracts}
\footnotesize
\setlength{\tabcolsep}{3pt}
\begin{tabular*}{\linewidth}{@{\extracolsep{\fill}}lccccc@{}}
\toprule
Task & Train & Dev. & Qpos & Action & RGB views \\
\midrule
StackCube & 900 & 110 & 9 & 8 & Base + wrist \\
TwoRobotStackCube & 900 & 143 & 18 & 16 & Base \\
G1TransportBox & 900 & 113 & 25 & 25 & Base + head \\
\bottomrule
\end{tabular*}
\end{table}

\paragraph{Sensor inputs and supervision.}
The actor receives two chronological frames of RGB images
and measured robot joint positions. At reset, the first frame is
duplicated. Native RGB is resized to $128\times128$ with bilinear
interpolation and normalized with the pretrained
ResNet18~\cite{he2016identity} image statistics. Joint histories are concatenated to dimensions 18, 36,
and 50 for StackCube, TwoRobotStackCube, and G1TransportBox.
Joint and action normalizers are fitted only on the 900 training
trajectories; action statistics use expert-valid labels.
Object poses, relative object positions, task state, contact flags,
forces, and privileged simulator state are excluded from actor inputs.
The expert mask excludes artificial failure actions from
supervised targets while retaining failure observations in histories.
Invalid and terminal-padding targets are zeroed before the
conditional variational autoencoder
(CVAE)~\cite{kim2021conditional} action embedding and
masked from posterior attention. The $L_1$ loss averages only
expert-valid action components; the Kullback--Leibler (KL) loss
averages only samples with at least one valid expert target.

\paragraph{Backbone and execution.}
Both methods use the same pretrained ResNet18 visual backbone and the transformer architecture
used by ACT~\cite{zhao2023learning}, with hidden width 256 and
feedforward width 512,
two encoder layers, four decoder layers, eight attention heads,
and a 32-dimensional CVAE latent.
Training predicts 16 native action vectors beginning at the current
observation time. During evaluation, we execute the first predicted
action and infer again at each 20-Hz control tick, without temporal
ensembling. This shared execution schedule was fixed after a separate
execution diagnostic and before evaluation on the fresh seed bank.
Inference uses a zero prior latent with the policy in evaluation mode.
Actions retain the native normalized \texttt{pd\_joint\_delta\_pos}
controller interface. For TwoRobotStackCube, the flat 16-dimensional
action preserves the left/right eight-component Panda blocks;
for G1TransportBox, all 25 components control native joint deltas.

\paragraph{Proprioceptive conditioner.}
In the main-text notation, $x_t\in\mathbb{R}^{d}$ is the normalized
two-frame joint history, with $d=18$, $36$, and $50$ for StackCube,
TwoRobotStackCube, and G1TransportBox, respectively.
The baseline uses an affine projection followed by
root mean square normalization (RMSNorm)~\cite{zhang2019root}.
For \sname{}, the degree-2 construction is
\begin{align}
 A_i(x_t)&=W_i x_t+b_i,\qquad i\in\{1,2\},
 \label{eq:appendix_act_affine} \\
 \psi_2(x_t)&=A_1(x_t)+s_2\bigl(A_1(x_t)\odot A_2(x_t)\bigr),
 \label{eq:appendix_act_polynomial} \\
 z_t&=g(\psi_2(x_t))
 =\operatorname{RMSNorm}\!\left(W_p\psi_2(x_t)+b_p\right).
 \label{eq:appendix_act_conditioner}
\end{align}
Here, $\odot$ denotes element-wise multiplication,
$W_i\in\mathbb{R}^{256\times d}$,
$b_i\in\mathbb{R}^{256}$,
$W_p\in\mathbb{R}^{256\times256}$, and
$b_p\in\mathbb{R}^{256}$.
Thus $A_i(x_t)$, $\psi_2(x_t)$, and $z_t$ are 256-dimensional.
Both conditioners use a trainable RMSNorm gain initialized to one
and a numerical stabilizer $\epsilon=10^{-6}$.
The scalar interaction scale $s_2$ is learned and initialized to $0.01$.
At initialization, $A_1$ copies the baseline affine projection,
$W_p=I_{256}$, and $b_p=0$, where $I_{256}$ is the identity matrix.
With these initial parameters, setting $s_2=0$ would reproduce the
baseline encoding; the actual initialization uses $s_2=0.01$.
The visual, latent, transformer, and action-head initialization
is shared across methods.

\begin{table}[!htbp]
\centering
\caption{Shared ACT~\cite{zhao2023learning} training and deployment settings.}
\label{tab:appendix_act_training_details}
\label{tab:appendix_manip_training_details}
\footnotesize
\setlength{\tabcolsep}{3pt}
\begin{tabular}{@{}ll@{}}
\toprule
Item & Setting \\
\midrule
RGB size / observation history & $128\times128$ / 2 frames \\
Hidden / feedforward width & 256 / 512 \\
Encoder / decoder / attention heads & 2 / 4 / 8 \\
CVAE latent / KL weight & 32 / 10 \\
Prediction queries / execution & 16 / first action at each 20-Hz tick \\
Temporal ensembling & Disabled \\
Batch / training seed / updates & 64 / 1000 / 100K \\
Optimizer & AdamW~\cite{prismapp_loshchilov2019adamw}, FP32 \\
Learning rate / backbone rate & $10^{-4}$ / $10^{-5}$ \\
Weight decay & $10^{-4}$ \\
Scheduler & StepLR at 66,666 updates, factor $0.1$ \\
Checkpoint & Final 100K EMA; configured power 0.75 \\
Conditioner width / RMSNorm epsilon & 256 / $10^{-6}$ for both methods \\
\bottomrule
\end{tabular}
\end{table}

\paragraph{Native evaluation.}
Each method and task uses five evaluation groups (2200--2204)
with 100 attempts each. Group index $r\in\{0,\ldots,4\}$ uses
reset seeds $1200000+1000r+i$, where $i\in\{0,\ldots,99\}$;
its displayed group ID is $2200+r$.
The seed bank was fixed before the reported evaluation outcomes.
Both methods share seeded native initial states and geometry
signatures. Success is native \texttt{info.success} at any executed
tick within 100 control steps (5 seconds); every scheduled attempt,
including a runtime error, remains in the denominator.
The success rates in Table~\ref{tab:dp_manipulation_success} are
mean $\pm$ sample standard deviation (SD) across these five groups,
using one trained checkpoint per method and task.
These SDs do not measure training-seed variability.
Training-geometry disjointness is not asserted because there is no
verified mapping from evaluation reset IDs to the demonstration geometry.

Evaluation uses ManiSkill~\cite{tao2025maniskill3} version 3.0.0b21,
with single-environment CPU PhysX and a Vulkan renderer bound to
the allocated NVIDIA GPU. The renderer is initialized before policy CUDA.
Both methods use this same backend, sensor preprocessing, and controller.
This compatibility path differs from CUDA-physics execution and is
not a numerical reproduction of an author-reported benchmark score.

\begin{table}[!htbp]
\centering
\caption{\textbf{\sname{} improves mean StackCube success under the shared evaluation protocol.}
ACT~\cite{zhao2023learning} and its \sname{} variant use the final
100K EMA checkpoint; each group contains 100 native
ManiSkill3~\cite{tao2025maniskill3} attempts.
Mean and SD are computed across evaluation groups, not training seeds.
The mean rises from $58.8\%$ to $68.8\%$ with the same reset-seed bank.}
\label{tab:appendix_act_seed_success}
\label{tab:appendix_libero_category_success}
\footnotesize
\setlength{\tabcolsep}{3pt}
\begin{tabular*}{\linewidth}{@{\extracolsep{\fill}}lcccccc@{}}
\toprule
Method & 2200 & 2201 & 2202 & 2203 & 2204 & Mean $\pm$ SD \\
\midrule
ACT & 62 & 61 & 50 & 55 & 66 & $58.8\pm6.3$ \\
\sname{} & 79 & 64 & 67 & 67 & 67 & $\mathbf{68.8\pm5.8}$ \\
\bottomrule
\end{tabular*}
\end{table}

\subsection{BFM-Zero and SmolVLA}
\label{sec:appendix_strong_backbones}

Table~\ref{tab:appendix_strong_backbone_setup} specifies the
architecture and evaluation settings for the reference-motion and
multi-task manipulation comparisons. The BFM-Zero~\cite{li2025bfm}
capacity control
widens the six-layer main network from 2048 to 2560 units; the
SmolVLA~\cite{shukor2025smolvla} control replaces the linear state projection with a three-layer MLP
of width 2048.
Tables~\ref{tab:bfm_zero_capacity} and~\ref{tab:smolvla_results}
count deployment-time BFM-Zero modules and the full SmolVLA
policy, respectively.

\paragraph{BFM-Zero reference-motion tracking.}
The history-state encoder uses the degree-2 representation,
a two-layer projection, and RMSNorm~\cite{zhang2019root},
while retaining the original
core network~\cite{li2025bfm}. These post-polynomial operations
constitute the backbone-specific mapping $g$.
All variants are trained from scratch with the native replay setup
and motion data. Nominal evaluation disables dynamics randomization,
sets static and dynamic terrain friction to $1.0$, and retains the
original link masses. We use the upstream tracking Earth Mover's
Distance (EMD) evaluator on the LAFAN motion set~\cite{harvey2020robust}.
The nominal and changed-dynamics comparisons use the same checkpoints
after 9.6M environment steps, with one training seed.
\paragraph{BFM-Zero encoder configuration.}
The encoder configuration uses the 29-degree-of-freedom G1 robot.
The observation configuration is
\texttt{humanoidverse/}\allowbreak\texttt{config/}\allowbreak
\texttt{obs/}\allowbreak\texttt{bfm\_zero\_obs.yaml}.
Its history stream stacks four frames of base angular velocity (3),
projected gravity (3), joint positions (29), joint velocities (29),
and previous actions (29), giving $4\times93=372$ inputs.
It specifies two 372-dimensional affine factors and an
\emph{element-wise} interaction scale:
\[
 \psi_2(x_t)=A_1(x_t)+s_2\odot A_1(x_t)\odot A_2(x_t),
 \qquad s_2\in\mathbb R^{372}.
\]
Each component of $s_2$ is learned and initialized to $0.01$.
The backbone-specific mapping $g$ is a $372\to372\to372$
projection with final RMSNorm~\cite{zhang2019root}
($\epsilon=10^{-6}$). The remaining observation streams retain their
original interfaces.
The scenario driver invokes \texttt{humanoidverse.tracking\_eval}
with the 29-degree-of-freedom LAFAN data, 128 evaluation environments,
and the fixed-friction/mass overrides for each scenario.
These dimensions describe the encoder configuration; the
reported training budget and evaluation conditions are summarized in
Table~\ref{tab:appendix_strong_backbone_setup}.

\paragraph{SmolVLA.}
We train on 40 LIBERO~\cite{liu2023libero} tasks using the
\texttt{HuggingFaceVLA/libero} release associated with
SmolVLA~\cite{shukor2025smolvla}.
The 16-layer visual-language backbone is initialized from
SmolVLM2-500M-Video-Instruct
and frozen; the action expert and
proprioceptive conditioner are trained from scratch.
The polynomial conditioner uses RMSNorm~\cite{zhang2019root} within $g$.
Training uses AdamW~\cite{kingma2014adam}
with learning rate $10^{-4}$, 100 warm-up
steps, and cosine decay over a 100K-update schedule.
The main comparison uses aligned 80K-update checkpoints and
reports mean $\pm$ SD over three training seeds.
Table~\ref{tab:appendix_smolvla_libero} instead gives a single-seed
suite breakdown; it is not the three-seed summary in
Table~\ref{tab:smolvla_results}.

\begin{table}[!htbp]
\centering
\caption{BFM-Zero~\cite{li2025bfm} and
SmolVLA~\cite{shukor2025smolvla} implementation settings.
The SmolVLA suite breakdown is a single run, distinct from
the main three-seed summary.}
\label{tab:appendix_strong_backbone_setup}
\footnotesize
\setlength{\tabcolsep}{4pt}
\renewcommand{\arraystretch}{1.08}
\begin{tabular}{@{}p{0.23\linewidth}p{0.35\linewidth}p{0.36\linewidth}@{}}
\toprule
Item & BFM-Zero & SmolVLA \\
\midrule
Data / benchmark & 40 LAFAN motions~\cite{harvey2020robust} & 40 LIBERO tasks~\cite{liu2023libero} \\
\sname{} input & History-state stream & Proprioception \\
Baseline architecture & Main network: width 2048, six layers & Linear state projection \\
Larger control & Main network: width 2560, six layers & State MLP: width 2048, three layers \\
Reported checkpoint & 9.6M environment steps & 80K updates \\
Batching & 512 parallel environments & Batch size 64 \\
Training seeds & Single seed: 1000 & Main: three seeds; breakdown: 1000 \\
Evaluation & 40 motions, 128 environments & 50 rollouts/task, 2000 total \\
\bottomrule
\end{tabular}
\end{table}

\begin{table}[!htbp]
\centering
\caption{\textbf{\sname{} achieves the highest mean success in the single-seed SmolVLA breakdown.}
SmolVLA~\cite{shukor2025smolvla} variants are evaluated on
LIBERO~\cite{liu2023libero} with seed 1000 and the protocol in
Table~\ref{tab:appendix_strong_backbone_setup}.
Avg. averages the four suites; bold marks each column's highest value.
\sname{} leads on three suites and the overall average,
while the larger model leads on Object.}
\label{tab:appendix_smolvla_libero}
\footnotesize
\setlength{\tabcolsep}{5pt}
\begin{tabular}{@{}lccccc@{}}
\toprule
Method & Spatial & Object & Goal & Long & Avg. \\
\midrule
SmolVLA & 69.8 & 57.2 & 81.2 & 45.8 & 63.50 \\
Larger Model & 64.6 & \textbf{65.6} & 85.0 & 44.4 & 64.90 \\
\sname{} & \textbf{70.0} & 57.4 & \textbf{85.4}
& \textbf{53.4} & \textbf{66.55} \\
\bottomrule
\end{tabular}
\end{table}

\section{Practical Benefits for Robot Control}
\label{sec:appendix_practical_control}
\label{sec:appendix_rq3} 

\subsection{Controlled failure recovery on StackCube}
\label{sec:appendix_recovery_protocol}
\label{sec:appendix_execution_perturbations}

Table~\ref{tab:libero_mcc_comparison} evaluates recovery from
44 shared physically induced drop states in
ManiSkill3 StackCube~\cite{tao2025maniskill3}.
ACT~\cite{zhao2023learning} and \sname{} use the same CroSTAta
recovery demonstrations~\cite{minelli2025crostata} and the final
100K checkpoints described in Sec.~\ref{sec:appendix_act_maniskill}.
This controlled-drop pilot is distinct from ordinary task success
and from an official CroSTAta recovery evaluation protocol.

\paragraph{Failure-state construction.}
A fixed baseline ACT checkpoint generates the setup trajectories.
When cubeA is grasped, lifted at least 3\,cm above its resting
height, and at least 8\,cm from cubeB in the horizontal plane
for three consecutive control ticks, we hold the arm and open
the gripper. A valid failure state is unfinished, ungrasped,
settled on the table, and inside the workspace.
We take the first 44 valid states from a fixed candidate order,
without selecting by either policy's recovery outcome.

Each ACT/\sname{} evaluation obtains its failure state by replaying
the same recorded setup actions from the same native reset,
rather than restoring a simulator snapshot.
We verify the physical state, controller targets, counters,
and two-frame sensor history before policy execution.
Floating physical quantities use absolute tolerance $10^{-6}$;
recorded actions, integer counters, and RGB history are compared exactly.
Object positions and forces are not provided to the ACT or \sname{} actor.

\paragraph{Recovery criteria and aggregation.}
After handoff, each policy has 100 control ticks (5 seconds) to recover,
with its history initialized from the last two setup observations
and prediction at every 20-Hz tick.
Recovery success is native StackCube completion after the drop.
Regrasp-and-lift requires grasping cubeA and lifting it by at least
3\,cm for three consecutive ticks; this is reported separately
and is not counted as task completion.
Every scheduled state remains in the denominator, including
handoff mismatches and runtime errors.

The time entry in Table~\ref{tab:libero_mcc_comparison} is averaged
over all 44 attempts. Successful attempts contribute their time
from handoff to first task completion; unsuccessful attempts
contribute the 5-second cap. Terminal placement error is also
averaged over all attempts, including failures.
Thus the time metric incorporates failed recoveries rather than
measuring speed only on successful trials.
The pilot reports proportions, not five-seed standard deviations.
Its baseline-generated failure distribution and small sample size
limit the scope of the comparison.
For StackCube, terminal placement error is
\[
 e_p=1000\left\lVert
 p_A-p_B-\begin{bmatrix}0&0&0.04\end{bmatrix}^{\!\top}
 \right\rVert_2\quad\text{(mm)},
\]
where $p_A$ and $p_B$ are the three-dimensional world-frame
centers of cubeA and cubeB at first completion or the 5-second cap.
The 4-cm offset is twice the native cube half-size of 2\,cm.
This is task-placement error, not end-effector command-tracking error.
The native stacking goal does not define a target cube yaw;
we therefore do not assign an orientation error to this goal.

\paragraph{Execution-time correction baselines.}
MCC-Sensorless and MCC-Oracle are simulation adaptations of
Minimalist Compliance Control~\cite{shi2026minimalist} applied to
a frozen ACT policy. MCC-Sensorless uses force-proxy corrections;
MCC-Oracle uses simulator contact-force measurements.
Both are evaluated on the same 44 validated states. The translation
correction acts on a Cartesian target derived from the native ACT
joint target, and is mapped back to the native joint-action interface
as described below.
\sname{} changes the learned representation rather than adding
an execution-time admittance controller, and receives no force,
wrench, or tactile feedback.
The sensorless wrapper estimates external torque as the negative
native PD-drive torque, rather than reading simulator contact force.
The drive model is
\[
 \tau_{\rm PD}=\operatorname{clip}
 \bigl(K_p(q^{\rm target}-q)-K_d\dot q,\,\pm\tau_{\rm lim}\bigr),
\]
using the native controller gains and saturation limits.
Here $q$, $\dot q$, and $q^{\rm target}$ are the measured arm-joint
positions, velocities, and current PD-drive targets; $K_p$ and $K_d$
are the native proportional and derivative gains.
The clipping operation uses the corresponding joint-torque limits
$\tau_{\rm lim}$ componentwise.
The robot links have gravity disabled in this simulation controller.
Five 100-Hz physics samples are averaged per 20-Hz control tick,
and an exponential moving average uses new-sample weight $0.1$.
Let $\bar\tau$ be this filtered PD-drive torque vector and
$J_p\in\mathbb R^{3\times d_{\rm arm}}$ the world-frame translational
Jacobian for the $d_{\rm arm}$ controlled arm joints.
The proxy force is
\[
 \hat f=(J_pJ_p^\top+10^{-3}I_3)^{-1}J_p(-\bar\tau),
\]
where $I_3$ is the $3\times3$ identity matrix.
Jacobians use central finite differences with a $10^{-3}$-rad step.
The oracle instead uses the previous control tick's mean net
hand--environment contact impulse divided by the physics interval.
This net force differs from the sum of contact magnitudes plotted
in Fig.~\ref{fig:force_compliance}.

The translational correction uses virtual mass $m_v=1$\,kg,
stiffness $K_v=100$\,N/m, damping $B_v=20$\,N\,s/m,
and control interval $\Delta t_{\mathrm{ctrl}}=0.05$\,s:
\begin{align*}
 \dot\delta_{k+1}
 &=\dot\delta_k+\frac{\Delta t_{\mathrm{ctrl}}}{m_v}
       (f_k-K_v\delta_k-B_v\dot\delta_k),\\
 \delta_{k+1}&=\delta_k+\Delta t_{\mathrm{ctrl}}\dot\delta_{k+1}.
\end{align*}
Here $k$ indexes 20-Hz control ticks, $\delta_k\in\mathbb R^3$
is the world-frame translation offset in meters, and $\dot\delta_k$
is its velocity. The input $f_k$ is the sensorless proxy or the oracle
net force. Both states start at zero. The offset norm is capped at
3\,cm, and outward radial velocity is removed when the cap is reached.
The corrected Cartesian target uses forward
kinematics~\cite{lynch2017modern} of the nominal ACT joint target
plus $\delta_k$, with its orientation unchanged.
Five damped least-squares inverse-kinematics iterations use the
six-dimensional pose error, damping $0.1$, and native joint limits.

Let $D\in\mathbb R^{d_{\mathrm{arm}}\times d_{\mathrm{arm}}}$
be the diagonal matrix of native arm-joint delta scales, and let
$a^{\mathrm{ACT}}$ be the arm portion of the normalized ACT action.
The conversions are
\[
 q^{\mathrm{nom}}=q+Da^{\mathrm{ACT}},\qquad
 a^{\mathrm{corr}}=D^{-1}(q^{\mathrm{corr}}-q),
\]
followed by componentwise clipping to $[-1,1]$.
The vector $q^{\mathrm{corr}}$ is the corrected joint target returned by
inverse kinematics; gripper commands are unchanged.
With zero force and offset, the wrapper preserves the nominal action.
These baselines are simulation adaptations, not reproductions of the
original hardware telemetry or policies trained with force inputs.

\subsection{Contact-aligned force measurements}
\label{sec:appendix_contact_measurements}
\label{sec:appendix_compliance}
\label{sec:appendix_contact_diagnostics}
\label{sec:appendix_mcc_diagnostics}

Fig.~\ref{fig:force_compliance} complements the aggregate recovery
results with an illustrative hand--cube contact diagnostic.
A contact image is shown on the left, followed by four force-only
panels for ACT~\cite{zhao2023learning}, the MCC-Sensorless and
MCC-Oracle adaptations~\cite{shi2026minimalist}, and \sname{}.
All traces start from the same validated induced-drop state
(failure-bank case 2) and use identical force and time axes.

\paragraph{Contact-load definition.}
Let $k$ index physics samples with duration
$\Delta t_k=0.01\,\mathrm{s}$.
Let $\mathcal C_k$ be the set of contact points between the hand
and either task cube, and let $J_{k,c}$ be the contact-impulse
vector at point $c\in\mathcal C_k$.
The plotted scalar load is
\begin{equation}
 F_k^{\mathrm{hand}}
 =\frac{1}{\Delta t_k}
  \sum_{c\in\mathcal C_k}\lVert J_{k,c}\rVert_2.
 \label{eq:appendix_hand_contact_load}
\end{equation}
Impulses are measured in N\,s and the resulting load in N.
Summing the impulse magnitudes retains opposing finger-force
contributions rather than cancelling them in a resultant vector.
For each method, $t=0$ is the first post-handoff 100-Hz sample
at which hand--cubeA force reaches 1\,N.
The alignment event uses cubeA, whereas the plotted load includes
contacts with either task cube.

\begin{table}[!htbp]
\centering
\caption{Contact-recording protocol for the illustrated
induced-drop recovery on ManiSkill3
StackCube~\cite{tao2025maniskill3} in Fig.~\ref{fig:force_compliance}.
The common failure bank is used for aggregate recovery statistics;
the figure is a single contact-aligned example, not a bank average.}
\label{tab:appendix_contact_details}
\label{tab:appendix_mcc_details}
\label{tab:appendix_probe_controls} 
\footnotesize
\setlength{\tabcolsep}{3pt}
\renewcommand{\arraystretch}{1.08}
\begin{tabular}{@{}p{0.30\linewidth}p{0.65\linewidth}@{}}
\toprule
Item & Setting \\
\midrule
Recovery evaluation & 44 shared validated drop states \\
Illustrated state & Failure-bank case 2 \\
Plotted contact pair & Hand--cubeA and hand--cubeB \\
Physics / policy rate & 100\,Hz / 20\,Hz \\
Force / impulse units & N / N\,s \\
Contact alignment & First post-handoff hand--cubeA force $\geq1$\,N \\
Trace window & 1\,s before and 5\,s after first contact \\
ACT / \sname{} actor inputs & RGB images and measured joint positions \\
Oracle wrapper feedback & Simulator contact force \\
\bottomrule
\end{tabular}
\end{table}

\paragraph{Visualization and interpretation.}
The curves show the measured one-second pre-contact prefix and
five seconds of actual policy execution after contact, without
smoothing, padding, or scene resets. Execution continues after
first task completion only for this illustration.
In contrast, each attempt in Table~\ref{tab:libero_mcc_comparison}
ends at first completion or the 5-second horizon measured from handoff.
The selected case was the first earlier ACT-failure/\sname{}-recovery
contrast with actual contact for both policies; it was not selected
for low contact force. Replay agreement and image/trace provenance
are retained with the figure.

In this example, \sname{} rapidly relaxes contact force and first
completes recovery 0.85 seconds after contact, without force input
to the actor or an explicit admittance controller.
This is consistent with the compliance-like behavior described in
the main text. The low-load phase includes object release, and the
continued rollout later manipulates the object again.
The example does not isolate sustained force-induced yielding or
estimate mechanical compliance from a force--displacement relation.

\subsection{SO-101 tape pickup}
\label{sec:appendix_so101}

Table~\ref{tab:so101_tape} compares ACT~\cite{zhao2023learning}
and \sname{} on the physical
SO-101 robot~\cite{knight2025standard}. Both policies are trained on the
same 50 demonstrations and evaluated in 50 hardware rollouts each,
with the same hardware protocol. A rollout succeeds when the tape roll
is lifted at least 5\,cm above the table and remains at that height for
3 seconds. ACT completes 34 of 50 attempts and \sname{} completes
47 of 50, corresponding to 68\% and 94\% success.

\paragraph{Execution measurements.}
Consistent with the main text, joint motion describes arm joint-space
travel during the first 3 seconds of execution. Joint tracking error
describes the rollout-averaged discrepancy between commanded and
measured arm-joint angles. These summaries are reported in degrees;
neither is an angular-speed measurement. More initial travel describes
an execution pattern rather than a separate task-success criterion.
The higher pickup success for \sname{} is accompanied by lower reported
mean tracking error, connecting improved physical task completion with
closer command following without attributing success to motion magnitude
alone.

\paragraph{Collection and reset workflow.}
The hardware collection utilities use a calibrated SO-101 leader to
teleoperate a calibrated SO-101 follower. Both policy launchers read
camera inputs and reset settings from a common experiment configuration.
A collection manifest specifies object-placement conditions, repetitions,
and shuffled order. Between recordings, the operator resets the tape
roll to the next listed condition and returns the arm to a saved ready
pose; the utility also supports automatic ready-pose motion.
Autonomous evaluation begins from that pose and uses episodic execution
with return to the initial position between attempts.

\section{Generalizing to Out-of-distribution Physical Conditions}
\label{sec:appendix_physical_generalization}
\label{sec:appendix_rq2} 

The dynamics evaluations use fixed trained checkpoints without
retraining. Humanoid-Gym and mjlab measure survival under their
respective protocols; BFM-Zero measures reference-motion tracking EMD.
These metrics and evaluation horizons are not pooled across backbones.

\subsection{Humanoid-Gym dynamics generalization}
\label{sec:appendix_hg_robustness}

The flat-terrain G1 study in
Humanoid-Gym~\cite{gu2024humanoid} (Fig.~\ref{fig:locomotion_dynamics})
uses a separate training and evaluation protocol from
Table~\ref{tab:locomotion_main}.
It compares \sname{} with a parameter-matched MLP and evaluates
60-second survival, rather than the 24-second survival in the
main policy-performance table.
Final checkpoints are evaluated without selection on the dynamics
sweeps. This comparison therefore evaluates the representation
together with its training recipe.
Table~\ref{tab:appendix_hg_robustness_protocol} lists the settings.

\begin{table}[!htbp]
\centering
\caption{
\textbf{Humanoid-Gym dynamics-generalization protocol.}
Experiments use Humanoid-Gym~\cite{gu2024humanoid}.
Parameter pairs list matched MLP / \sname{}; actor--critic
totals include action-distribution parameters.
Sensor-noise scales are ordered as joint position, joint
velocity, angular velocity, and orientation, in rad,
rad/s, rad/s, and rad, respectively.
$\mathcal{U}[a,b]$ denotes a uniform distribution on $[a,b]$.
}
\label{tab:appendix_hg_robustness_protocol}
\footnotesize
\setlength{\tabcolsep}{3pt}
\renewcommand{\arraystretch}{1.12}
\begin{tabular}{@{}p{0.28\linewidth}p{0.31\linewidth}p{0.35\linewidth}@{}}
\toprule
Setting & Training & Evaluation \\
\midrule
PPO~\cite{schulman2017proximal} budget & 3,001 updates; 4,096 environments & Final checkpoint \\
Actor-mean parameters & 922,092 / 921,996 & Fixed \\
Actor--critic total parameters & 1,320,953 / 1,320,857 & Fixed \\
History / action dimension & $15\times47$ / 12 & Unchanged \\
Polynomial degree / warm-up & 2 / 500 updates, $\beta_{\mathrm{poly}}:0\rightarrow1$ & $\beta_{\mathrm{poly}}=1$ \\
\midrule
Effective friction $\mu$ & $\mathcal{U}[0.2,1.2]$ & $\{0.2,0.4,0.6,0.9,1.2,1.5,1.8\}$ \\
Ground / shape friction & $0.2$ / $\mathcal{U}[0.2,2.2]$ & Both set to $\mu$ \\
Robot-mass multiplier & $0.90,0.91,\ldots,1.10$ & $\{0.8,0.9,1.0,1.1,1.2\}$ \\
Backpack mass (kg) & $\{0,1.2,2.4\}$ & $\{0,1.2,2.4,3.6,4.8\}$ \\
XY replacement bound $b$ (m/s) & 0.2 & $\{0,0.1,0.2,0.4,0.6\}$ \\
Angular replacement bound (rad/s) & 0.4 & Disabled \\
Replacement interval & 4 s & 4 s for $b>0$ \\
Sensor-noise SD & $(0.03,0.30,0.06,0.018)$ & Disabled \\
Previous-action blend weight & $\mathcal{U}[0,0.5]$ & Disabled \\
Multiplicative action-noise SD & 0.02 & Disabled \\
\midrule
Forward command (m/s) & $\mathcal{U}[-0.3,0.6]$ & Unchanged \\
Lateral command (m/s) & $\mathcal{U}[-0.3,0.3]$ & Unchanged \\
Target heading (rad) & $\mathcal{U}[-3.14,3.14]$ & Unchanged \\
Heading gain / yaw-rate clip & $0.5$ / $[-1,1]$ rad/s & Unchanged \\
Planar-command dead zone & Norm $\leq0.2$ m/s & Unchanged \\
Command resampling & 8 s & 8 s \\
Episode horizon & 24 s & 60 s \\
Training / evaluation seeds & 1 training seed (seed 1) & 5 evaluation seeds (301--305) \\
Episode quota & --- & 2 per environment; 100 environments per seed \\
\bottomrule
\end{tabular}

\end{table}

\paragraph{Training distribution.}
The simulator combines contact friction as
$\mu_{\mathrm{eff}}=(\mu_{\mathrm{ground}}+\mu_{\mathrm{robot}})/2$,
where the subscripts denote effective, ground, and robot-shape friction.
Discrete robot-mass and backpack settings are assigned through
balanced, shuffled Cartesian-product combinations shared across methods.
Link scaling changes the original robot links' masses and inertias,
excluding the separately attached backpack, which has its own mass,
center of mass, and inertia.
Sensor noise is applied before history stacking, excluding command
and previous-action channels. The action-blending weight is resampled
at reset. The scheduled branch scale $\beta_{\mathrm{poly}}$ in
Eq.~\ref{eq:appendix_hg_encoder} increases from zero to one over
500 updates and is fixed to one during evaluation. This schedule is
distinct from the learned ACT interaction scale $s_2$.

\paragraph{Evaluation sweeps.}
Each sweep changes one factor from nominal effective friction
$\mu=0.6$, original link mass, no backpack, and no root-velocity
perturbation. The nominal evaluation is reused across panels.
Out-of-range settings refer to the individually varied training
supports; a training boundary is not an unseen value.
The six out-of-range conditions summarized in the main text comprise
friction values $\{1.5,1.8\}$, link-mass multipliers $\{0.8,1.2\}$,
and backpack masses $\{3.6,4.8\}$\,kg, varied one factor at a time.

The additional root-velocity sweep in the protocol replaces each
world-frame XY velocity component with an independent draw from
$\mathcal{U}[-b,b]$ on the native schedule, including reset-time
scheduling; $b$ is the replacement bound and $b=0$ disables replacement.
This is neither an applied force nor an additive velocity increment.
Matching the training XY bound does not reproduce the full training
distribution, since angular perturbations and observation/action
noise are disabled during evaluation.
Fig.~\ref{fig:locomotion_dynamics} displays the friction, mass,
and payload sweeps, not this additional velocity-replacement sweep.

\paragraph{Commands and survival.}
Yaw-rate commands are computed from heading error using the gain and
clipping in Table~\ref{tab:appendix_hg_robustness_protocol}, rather
than sampled directly. Native resets and the initial zero-action
step are retained. Survival uses the native timeout flag, including
coincident timeout and physical-termination events, and does not
impose a velocity-tracking threshold.
The Humanoid-Gym panels report mean and sample SD over five
evaluation seeds, not independently trained policies.
Each seed uses 100 environments and two episodes per environment.
Methods share evaluation settings and seeds, but their actions
and reset times can differ.

\subsection{mjlab rough-terrain survival}
\label{sec:appendix_mjlab_robustness}

The lower row of Fig.~\ref{fig:locomotion_dynamics} compares
\sname{} with a standard MLP in mjlab~\cite{zakka2026mjlab}
on random rough terrain, reporting 20-second supported survival.
Friction, robot mass, and payload are varied one factor at a time,
using fixed trained checkpoints without retraining.
This is a separate protocol from the 60-second flat-terrain
Humanoid-Gym evaluation.
Across the six out-of-range conditions summarized in the main text,
mean survival is 87.1\% for \sname{} and 80.0\% for the MLP.
Those conditions span friction coefficients of 0.2--1.5,
link-mass scaling factors of 0.6--1.4, and payloads of 6--8\,kg.
\paragraph{Training configuration.}
Each method is trained for 30,000
PPO~\cite{schulman2017proximal} updates with seed 1,
4,096 environments, and 24 control samples per update.
Evaluation uses \texttt{model\_29999.pt}.
The physics interval is 0.005\,s
with decimation four, giving a 50-Hz policy and 20-second episodes.
The actor and critic use ELU~\cite{clevert2016fast} MLPs
with hidden widths $[512,256,128]$;
the actor has RMSNorm~\cite{zhang2019root} ($\epsilon=10^{-6}$) after its final hidden
activation and before its action projection for both methods.
The actor receives 286 values: 99 robot/command values and 187
terrain-height samples; the critic receives 298 and there are 29 actions.
In the main-text partition, $x_t\in\mathbb R^{99}$ is the
robot/command stream and the terrain-height samples are part of $c_t$.
Empirical observation normalization is shared across methods.
\sname{} transforms the 99 robot/command values with two
$99\to256$ affine factors, a learned element-wise interaction scale $s_2\in\mathbb R^{256}$
initialized componentwise to $0.01$, and a $256\to99$ projection. The 187 height samples
retain their original path into the actor.
PPO uses Adam~\cite{kingma2014adam}
with initial learning rate $10^{-3}$, adaptive
KL target $0.01$, clipping $0.2$, five epochs, four minibatches,
$\gamma=0.99$, entropy weight $0.01$,
and maximum gradient norm $1$.

Training uses the native mixed-terrain curriculum: flat terrain,
ascending stairs, and descending stairs each have proportion $0.2$;
ascending slopes, descending slopes, random rough terrain, and
waves each have proportion $0.1$.
Terrain patches are $8\times8$\,m on a $10\times20$ grid,
with a 20-m border and initial terrain level at most five.
The generator's difficulty range is $[0,1]$.
The random-rough component has height-noise range $[0.02,0.10]$\,m,
noise step $0.02$\,m, horizontal scale $0.1$\,m, and vertical
scale $0.005$\,m. The figure evaluates random rough terrain only.

At startup, a shared nominal mask selects friction $0.6$,
mass/inertia scale $1$, and zero payload with probability $0.2$.
Otherwise, friction, link-density scale, and payload are sampled
independently from $\mathcal U[0.3,1.2]$,
$\mathcal U[0.8,1.2]$, and $\mathcal U[0,5]$\,kg.
Density scales original link mass and inertia together; the
original robot mass is 33.341142\,kg, excluding the payload.
The payload is fixed to the torso at offset $(-0.15,0,0.20)$\,m,
with box dimensions $(0.20,0.12,0.16)$\,m and no added collisions.
Native center-of-mass, encoder, and push randomization remain active,
including for the nominal-mask samples.
The forward-command curriculum uses $[-1,1]$\,m/s initially,
$[-1.5,2]$ after 120K control steps, and $[-2,3]$ after 240K.

\paragraph{Evaluation geometry and survival.}
Five paired layout seeds, 401--405, each provide 200 native resets,
giving 1,000 attempts per method and condition.
Episode seeds are $1000s+i$, with layout seed $s$ and
$i\in\{0,\ldots,199\}$.
Each layout uses a common certified $21\times21$ grid of
$8\times8$-m patches (bounds $\pm84$\,m), curriculum disabled,
and fixed difficulty $0.5$, retaining the native rough-terrain
noise, platforms, and borders.
Physical sweeps use command $(v_x,v_y,\omega_z)=(0.5,0,0)$
from rest, without a warm-up interval.
Supported survival requires the full 20 seconds without native
fall/termination and with both the root XY position and all 14
projected foot capsules inside the certified grid after every
control step, including the final one.
No velocity-tracking threshold is imposed, and failed attempts
remain in the denominator.
Error bars are pointwise Wilson 95\% intervals~\cite{prismapp_wilson1927interval}
for the pooled
success count out of 1,000, conditional on these fixed layouts
and checkpoints; they are not training-seed standard deviations.

The six displayed out-of-range settings are friction $0.2$ and
$1.5$, mass/inertia scales $0.6$ and $1.4$, and payloads 6 and
8\,kg. The separately measured friction-$0.1$ condition is omitted
from the displayed figure and this six-condition mean.

\subsection{BFM-Zero dynamics changes}
\label{sec:appendix_bfm_robustness}

Table~\ref{tab:bfm_zero_robustness} uses the same 9.6M-step
BFM-Zero~\cite{li2025bfm} checkpoints and tracking EMD evaluator
as the nominal comparison
in Table~\ref{tab:bfm_zero_capacity}.
The low-friction condition sets both static and dynamic terrain
friction coefficients to $0.20$; the robot-mass condition multiplies
link masses by $1.15$. Each changes one factor from nominal.
Link-mass scaling is not an attached payload.
All other scenario and rollout settings are retained.
Mean is the arithmetic average of the nominal, low-friction,
and increased-mass EMD values.
Both perturbations increase EMD for every model, but \sname{}
retains lower EMD than the original and larger-capacity baselines.

\section{Learned Interactions and Action Prediction}
\label{sec:appendix_rq4}

\subsection{Physical-response probing}
\label{sec:appendix_physical_probes}

Table~\ref{tab:physics_probe_by_domain} evaluates whether mechanics-inspired
response targets can be predicted linearly from frozen policy features.
The four target names and formulas follow the main table: slip velocity,
joint power, contact impulse, and contact work. Their simulator-derived
targets are used only for analysis, not as policy inputs or auxiliary
training objectives. Linear readouts are fitted on a fitting partition
and evaluated on held-out windows with horizon $H=5$.

\paragraph{Slip and joint-power targets.}
For Humanoid-Gym, $h\in\{1,\ldots,H\}$ indexes future control samples
after the window start $t$. The 100-Hz control rate gives a sample
interval of $0.01$\,s. Let $\mathcal F$ contain the two feet, and define
\[
 c_{f,k}=\mathbf 1\!\left[F^z_{f,k}>5\,\mathrm N\right],
\]
where $F^z_{f,k}$ is the vertical contact force for foot $f$ at sample
$k$, and $\mathbf 1[\cdot]$ is the indicator function.
The vector $v^{\mathrm{world}}_{f,xy,k}\in\mathbb R^2$ is that foot's
planar world-frame velocity. With joint torque $\tau_{j,k}$ and joint
angular velocity $\dot q_{j,k}$, the targets are
\begin{align*}
 y_t^{\mathrm{slip}}
 &=\sum_{h=1}^{H}\sum_{f\in\mathcal F}
       c_{f,t+h}\|v^{\mathrm{world}}_{f,xy,t+h}\|_2,\\
 y_t^{\mathrm{power}}
 &=\log\!\left(1+\sum_{h=1}^{H}\sum_{j=1}^{12}
                    |\tau_{j,t+h}\dot q_{j,t+h}|\right).
\end{align*}
These are the contact-masked slip and per-joint absolute-power formulas
in Table~\ref{tab:physics_probe_by_domain}.
Taking absolute values before summing prevents cancellation between
positive and negative joint-power contributions. Neither expression
contains a time-integration multiplier: they are sampled response
proxies, not integrated displacement or energy.

\paragraph{Contact-response targets.}
The two contact rows retain the discrete expressions reported in
Table~\ref{tab:physics_probe_by_domain}:
\begin{align*}
 y_t^{\mathrm{impulse}}
 &=\log\!\left(1+\sum_{h=1}^{H}F_{t+h}\right),\\
 y_t^{\mathrm{work}}
 &=\sum_{h=1}^{H}F_{t+h}\|\Delta x_{t+h}\|_2.
\end{align*}
Here $F_k$ denotes the scalar contact signal entering the target and
$\Delta x_k$ denotes the spatial displacement used in the contact-work
proxy. In these expressions, $\Delta x_k$ is not a difference between
policy-input vectors $x_t$. The work target combines a contact scalar
with a displacement magnitude rather than a signed force--displacement
inner product. The impulse row is retained in its reported discrete
form, without inserting an additional time-step multiplier.
Consequently, these labels describe mechanics-inspired target proxies,
not independently calibrated continuous-time impulse or work estimates.
The locomotion collection and readout settings below apply to the slip
and joint-power rows; they do not specify the contact-probe collection.

\paragraph{Locomotion feature collection.}
Both frozen locomotion policies are evaluated on a common observation
bank generated by the \sname{} policy. Collection disables observation
noise, action noise, and action delay, while retaining friction and
base-mass randomization and one-second pushes with XY bound
$0.2$\,m/s and angular bound $0.4$\,rad/s.
A window can start only when root height exceeds $0.55$\,m, absolute
roll and pitch are below $0.35$\,rad, episode age is at least ten control
ticks, and no reset flag is present. Selected windows contain a future
push with nonzero XY magnitude.
The MLP readout uses the final 128-dimensional hidden activation;
the \sname{} readout uses the 256-dimensional encoder output $z_t$
in Eq.~\ref{eq:appendix_hg_encoder}.

\paragraph{Fitting and held-out evaluation.}
Locomotion windows are randomly partitioned 80/20 into fitting and
evaluation subsets. Features and targets are standardized using only
fitting-set means and standard deviations, with $10^{-6}$ added to
each standard-deviation denominator. For standardized feature
$\widetilde\phi_n\in\mathbb R^{d_\phi}$ and scalar target
$\widetilde y_n$, the prediction is
$\widehat y_n=w^\top\widetilde\phi_n+b$, where
$w\in\mathbb R^{d_\phi}$ and $b\in\mathbb R$.
The weights use ridge regularization and the intercept is unpenalized;
the ridge strength and base split seed are run parameters.
The split seed is $s+100i$, where $s$ is the base seed and $i$ is the
method index, so the two methods need not use identical held-out subsets.
Splitting is at window level rather than trajectory level. This protocol
therefore measures held-out-window prediction, not generalization to
unseen trajectories or a matched-width/layer probe control.

\paragraph{Metrics and interpretation.}
The locomotion readouts report mean squared error (MSE) on standardized
targets and Pearson correlation coefficient (PCC) on held-out windows.
For the relative-change columns of Table~\ref{tab:physics_probe_by_domain},
the reported percentages are
\[
 \Delta_{\mathrm{MSE}}=
 100\frac{\mathrm{MSE}_{\mathrm{PRISM}}-\mathrm{MSE}_{\mathrm{base}}}
          {\mathrm{MSE}_{\mathrm{base}}},\qquad
 \Delta_{\mathrm{PCC}}=
 100\frac{\mathrm{PCC}_{\mathrm{PRISM}}-\mathrm{PCC}_{\mathrm{base}}}
          {\mathrm{PCC}_{\mathrm{base}}}.
\]
Here base denotes the corresponding baseline column.
Negative MSE changes and positive PCC changes indicate improvement.
The main table reports both improvements for each of its four targets.
This supports greater accessibility to the tested linear readouts;
it does not by itself identify a physical mechanism or show that a
policy uses the decoded quantity to generate actions.
The diagnostic in Sec.~\ref{sec:appendix_action_sensitivity}
addresses action dependence directly.

\subsection{Action sensitivity to multiplicative features}
\label{sec:appendix_action_sensitivity}

Table~\ref{tab:prism_learned_terms} measures how predicted actions change
when a complete latent product is removed from a frozen policy.
The domain-specific statistics are the same as those in the main caption:
mean joint-target $L_2$ distance for
Humanoid-Gym~\cite{gu2024humanoid} and coordinate-wise
mean absolute change in the first normalized action for
ACT~\cite{zhao2023learning}.

\paragraph{Latent-product intervention.}
Let $r_t\in\mathbb R^m$ be the multiplicative contribution at the
intervention location. Removing coordinate $j$ gives
\begin{equation}
 r_t^{(-j)}=r_t-[r_t]_j e_j,
 \label{eq:appendix_feature_ablation}
\end{equation}
where $[r_t]_j$ is the $j$th component and $e_j\in\mathbb R^m$ is
the corresponding coordinate vector.
For ACT, $r_t=s_2(A_1(x_t)\odot A_2(x_t))$ and
$\psi_2^{(-j)}(x_t)=A_1(x_t)+r_t^{(-j)}$ is passed through the
unchanged projection and RMSNorm in $g$.
For Humanoid-Gym, Eq.~\ref{eq:appendix_hg_product} uses
$r_t=A_1^{\mathrm{HG}}(x_t)\odot A_2^{\mathrm{HG}}(x_t)$ and
$\psi_2^{\mathrm{HG},(-j)}(x_t)=p_{0,t}+r_t^{(-j)}$.
Its affine base $p_{0,t}$, raw-input path $\mathcal R(x_t)$,
projection $\mathcal P$, and remaining actor are retained.
In both cases, the observations, learned weights, and all other latent
products are fixed; the intervention removes one product coordinate,
not its original input channels.

\paragraph{Humanoid-Gym statistic and evaluated candidates.}
Let $\widehat a_n$ and $\widehat a_n^{(-j)}$ be the original and
intervened native action vectors for saved observation $n$.
The diagnostic uses $N_{\mathrm{HG}}=512$ observations and a seed-2
checkpoint. After the native $0.25$-rad joint-target scaling, it reports
\[
 \Delta a_{\mathrm{HG}}^{(j)}=
 \frac{1}{N_{\mathrm{HG}}}\sum_{n=1}^{N_{\mathrm{HG}}}
 \left\|0.25\bigl(\widehat a_n-\widehat a_n^{(-j)}\bigr)\right\|_2.
\]
The action-vector distance is in radians, not a coordinate-averaged
absolute error. The evaluated candidate coordinates are 159, 194, 19,
and 154. The displayed humanoid entries are selected within this
candidate set; their ordering does not assert an exhaustive ranking
of all 256 products. The main table reports action changes of
$0.057$ and $0.025$ for its two displayed humanoid interactions.

\paragraph{ACT statistic and feature ranking.}
For each manipulation task, the analysis uses its final 100K EMA
checkpoint with training seed 1000 and $N_{\mathrm{ACT}}=128$ held-out
RGB--joint-history windows. Four time-stratified observations are drawn
from each of 32 development trajectories using sampling seed 20261016,
without success-based filtering. Each of the 256 products is ablated
separately before the fixed projection and RMSNorm.
For action dimension $d_a$, the statistic is
\[
 \Delta a_{\mathrm{ACT}}^{(j)}=
 \frac{1}{N_{\mathrm{ACT}}d_a}
 \sum_{n=1}^{N_{\mathrm{ACT}}}\sum_{\ell=1}^{d_a}
 \left|\widehat a_{n,\ell}-\widehat a_{n,\ell}^{(-j)}\right|.
\]
The action vectors are the first predicted normalized actions, before
denormalization or clipping. Products are ranked within each task by
this statistic. Table~\ref{tab:prism_learned_terms} displays the two
highest-ranked TwoRobotStackCube products, with changes of $0.00338$
and $0.00270$; it does not display a pair for every analyzed task.
The unablated and intervened outputs are computed with the same frozen
policy operations. Larger values indicate greater output sensitivity,
not better task performance.

\begin{table}[!htbp]
\centering
\caption{\textbf{Action-sensitivity statistics follow the domain-specific
 definitions in the main text.}
 The Humanoid-Gym~\cite{gu2024humanoid} and
 ACT~\cite{zhao2023learning} diagnostics remove one complete latent
 product and retain the other policy operations. Humanoid-Gym measures joint-target vector
 distance in radians; ACT measures coordinate-wise absolute change in
 normalized action units. Their magnitudes are therefore not directly
 comparable across domains.}
\label{tab:appendix_action_sensitivity_protocol}
\footnotesize
\setlength{\tabcolsep}{4pt}
\renewcommand{\arraystretch}{1.08}
\begin{tabular}{@{}p{0.24\linewidth}p{0.34\linewidth}p{0.35\linewidth}@{}}
\toprule
Item & Humanoid-Gym & ACT \\
\midrule
Observation windows & 512 saved observations & 128 held-out windows per task \\
Training checkpoint & Seed 2 & Final 100K EMA, seed 1000 \\
Action statistic & Mean $L_2$ distance & Coordinate-wise mean absolute change \\
Action representation & After $0.25$-rad target scaling & First normalized action, before clipping \\
Feature selection & Ranked within selected candidates & All 256 products ranked within task \\
\bottomrule
\end{tabular}
\end{table}

\paragraph{Interaction labels.}
Each displayed product names dominant input coefficients in two dense
affine factors; it is not an isolated raw-input monomial.
For Humanoid-Gym, let $W_{\mathcal H}$, $W_{\mathcal B_0}$,
$W_{\mathcal B_1}$, and $W_{\mathcal S_0}$ be the weight matrices
of the corresponding affine maps. The effective raw-input factor
matrices are $W_{\mathcal B_1}W_{\mathcal H}$ and
$W_{\mathcal S_0}W_{\mathcal B_0}W_{\mathcal H}$.
The label uses the largest-magnitude coefficient in the first factor
and the largest coefficient in the second factor excluding the first
selected input coordinate. For ACT, the label independently selects
the largest-magnitude coefficient in each affine row on normalized
joint histories; repeated coordinates are allowed.

The symbols $\dot q_{\mathrm{Lhippitch}}$ and
$\dot q_{\mathrm{Rhippitch}}$ denote left and right hip-pitch velocities.
In TwoRobotStackCube, $q_{\mathrm{L5}}$ and $q_{\mathrm{R7}}$ denote
the measured angles of the fifth joint of the left seven-degree-of-freedom
Panda arm and the seventh joint of the right arm. The symbol
$g_{\mathrm{R2}}$ denotes measured displacement of the right gripper's
second finger joint and is distinct from the feature mapping $g$.
The index $t-1$ denotes the preceding observation in the input history.
With the observations and affine paths unchanged, removing these
products alters actions in both tasks, showing that the learned
multiplicative features contribute to action prediction.
This intervention does not isolate the physical role of a single
raw-input product or establish a task-success benefit from it alone.

\subsection{Polynomial degree}
\label{sec:appendix_degree}

Table~\ref{tab:degree_ablation_humanoid} compares affine (D1), quadratic
(D2), and cubic (D3) constructions in the 256-wide Humanoid-Gym
polynomial-actor family. As in the main method, degree is the maximum
polynomial degree before the downstream nonlinear mapping, not the
degree of the full policy. These are separate ablation runs, not the
five-training-seed summary in Table~\ref{tab:locomotion_main}.

Each degree increment adds two biased $256\to256$ affine maps,
or 131,584 actor parameters. Adding the shared 398,849-parameter critic
and 12 action standard deviations gives total counts of 1,189,273,
1,320,857, and 1,452,441 for D1, D2, and D3, respectively.
Thus the comparison varies both degree and capacity.
Most of the observed improvement occurs from D1 to D2, with a smaller
additional improvement for D3; the quadratic construction remains the
default for its simpler form.

\begin{table}[!htbp]
\centering
\caption{\textbf{Quadratic features capture most of the observed polynomial-degree gain.}
D1, D2, and D3 denote affine, quadratic, and cubic feature constructions
on Humanoid-Gym~\cite{gu2024humanoid}.
These runs are separate from the five-seed comparison
in Table~\ref{tab:locomotion_main}; bold marks the best reported value.
Survival rises from $55.75\%$ for D1 to $91.00\%$ for D2,
with a further increase to $92.50\%$ for D3.
Actor capacity also increases with degree, so this is not a
parameter-matched isolation of polynomial degree.}
\label{tab:degree_ablation_humanoid}
\footnotesize
\setlength{\tabcolsep}{3pt}
\begin{tabular}{@{}lrcccc@{}}
\toprule
Method & Actor params & Return $\uparrow$ & Ep. Len. $\uparrow$
& Lin. Err. $\downarrow$ & Survival (\%) $\uparrow$ \\
\midrule
\sname{}-D1 & 790,412 & 75.26 & 1408.4 & 0.4685 & 55.75 \\
\sname{}-D2 & 921,996 & 131.78 & 2198.7 & 0.2254 & 91.00 \\
\sname{}-D3 & 1,053,580 & \textbf{134.60} & \textbf{2296.5}
& \textbf{0.2097} & \textbf{92.50} \\
\bottomrule
\end{tabular}
\end{table}

\clearpage


\end{document}